\documentclass[10pt]{article}
\usepackage[a4paper]{geometry}
\usepackage[utf8]{inputenc}
\usepackage{algorithm2e}
\usepackage{amsmath}
\usepackage{mathtools}
\usepackage{amssymb}
\usepackage{tocloft}
\usepackage[export]{adjustbox}
\usepackage[dvipsnames]{xcolor}
\usepackage{blindtext}
\usepackage{graphicx}
\usepackage{float}
\usepackage{parskip}
\usepackage{fancyhdr}
\usepackage{tabularray}
\usepackage{wrapfig}
\usepackage{csquotes}

\usepackage{bm}

\usepackage{natbib}
\setcitestyle{square}
\setcitestyle{citesep={,}}

\numberwithin{equation}{section}

\usepackage{tikz}
\usetikzlibrary{bayesnet}
\usetikzlibrary{arrows}
\usetikzlibrary{arrows.meta}
\usetikzlibrary{backgrounds}
\usetikzlibrary{positioning}
\pgfdeclarelayer{cap}
\pgfsetlayers{main,cap}

\usepackage{caption}
\usepackage{subcaption}

\usepackage{titling}
\predate{}
\date{}
\postdate{\vspace{\baselineskip}}

\usepackage{enumitem}
\setlist[itemize]{align=parleft,left=0pt..1em}

\usepackage{abstract}

\usepackage{setspace}

\usepackage{titlesec}
\titleformat{\section}{\normalfont\Large}{\thesection}{1em}{}
\titleformat{\subsection}{\normalfont\large}{\thesubsection}{1em}{}
\titleformat{\subsubsection}{\normalfont\normalsize}{\thesubsubsection}{1em}{}
\titlespacing{\paragraph}{0pt}{0.5\baselineskip+2pt}{1em}

\usepackage[linktocpage]{hyperref}
\hypersetup{
	breaklinks=True,
    colorlinks=True,
	linkcolor=blue,    
    urlcolor=gray,
    citecolor=ForestGreen,
}

\newcommand{\mbb}[1]{\mathbb{#1}}
\newcommand{\mbf}[1]{\mathbf{#1}}
\newcommand{\mcal}[1]{\mathcal{#1}}
\newcommand{\msf}[1]{\mathsf{#1}}
\newcommand{\bsb}[1]{\boldsymbol{#1}}
\newcommand{\tbf}[1]{\textbf{#1}}
\newcommand{\tpm}{\textpm}

\title{
	\rule{\textwidth}{3pt}\vspace{5pt}
	\LARGE{\tbf{Amortized Variational Inference \\ for Deep Gaussian Processes}}\vspace{-2.28pt}
	\rule{\textwidth}{1pt}\vspace{\baselineskip}
}
\author{
	\normalsize\textbf{Qiuxian Meng} \\
	\normalsize Xiamen University \\
	\normalsize\href{mailto:qxmengxmu@outlook.com}{qxmengxmu@outlook.com}
	\and
	\normalsize\textbf{Yongyou Zhang} \\
	\normalsize Xiamen University \\
	\normalsize\href{mailto:yongyouzhang@xmu.edu.cn}{yongyouzhang@xmu.edu.cn}
}

\begin{document}

\newgeometry{left=3cm, right=3cm, top=3cm, bottom=3cm, headsep=0.5cm, footskip=1cm}
\captionsetup{width=\textwidth-2cm}

\maketitle

\pagestyle{fancy}
\fancyhead{}
\fancyhead[C]{\tbf{Amortized Variational Inference for Deep Gaussian Processes}}

\begin{center}
	\parbox[c]{\textwidth-2cm}{
		\textbf{Abstract}: Gaussian processes (GPs) are Bayesian nonparametric models
		for function approximation with principled predictive uncertainty estimates.
		Deep Gaussian processes (DGPs) are multilayer generalizations of GPs
		that can represent complex marginal densities as well as complex mappings.
		As exact inference is either computationally prohibitive
		or analytically intractable in GPs and extensions thereof,
		some existing methods resort to variational inference (VI) techniques for tractable approximations.
		However, the expressivity of conventional approximate GP models critically relies on
		independent inducing variables that might not be informative enough for some problems.
		In this work we introduce amortized variational inference for DGPs, which learns
		an inference function that maps each observation to variational parameters.
		The resulting method enjoys a more expressive prior conditioned on fewer input dependent inducing variables
		and a flexible amortized marginal posterior that is able to model more complicated functions.
		We show with theoretical reasoning and experimental results that our method performs similarly
		or better than previous approaches at less computational cost.
	}
\end{center}
\vspace{\baselineskip}

\section{Introduction}

Gaussian processes (GPs, \citet{10.7551/mitpress/3206.001.0001}) are effective models
for black-box function approximation with wide adoption in machine learning (ML)
for both supervised and unsupervised tasks \citep{Lawrence2003GaussianPL}.
The Bayesian nonparametric nature empowers GPs with attractive properties:
GPs are unsusceptible to overfitting while offering principled predictive uncertainty estimates
that are critical in some cases; GPs can represent a rich class of functions
with few hyperparameters as the model complexity grows with data.
However, GPs often struggle to model complex processes in practice,
since the expressivity is limited by the kernel/covariance function.
This limitation has motivated the introduction of Deep Gaussian processes
(DGPs, \citet{pmlr-v31-damianou13a}), which stack multiple layers of GPs hierarchically
to build up more flexible function priors while preserving the intrinsic capabilities.

Exact inference in GPs and the extensions thereof is hindered by cubic computational complexity
\(\mcal{O}(N^3)\) and quadratic memory complexity \(\mcal{O}(N^2)\) in sample size,
then one has to resort to approximate inference, of which the most popular being
sparse GPs with stochastic variational inference (SVI, \citet{10.5555/3023638.3023667}).
Such methods replaced the exact prior with an approximate GP prior wherein \(M\ll N\) inducing points
are treated as variational parameters and jointly optimized together with a variational posterior.
The computational cost of this approach scales down to \(\mcal{O}(M^3)\) that is free of \(N\),
making GPs scalable to very large datasets.
\citet{NIPS2017_82089746} proposed a doubly stochastic variational inference algorithm for DGPs with
sparse GP layers that induces an approximate posterior composed of layer-wise variational posterior marginals.

Conventional sparse GPs heavily rely on inducing points to create correct approximations to the target function.
One often observes that after convergence the inducing points are located in those regions of the input space
in which the latent function changes intensively.
Therefore a large number \(M\) of inducing points are required to model complex functions,
making it expensive for sparse GPs to get good predictions in those problems.
The scaling issue is even worse for DGPs, which use multioutput GPs as intermediate layers.

There have been several efforts to further improve the expressivity and scalability of sparse GPs in the literature,
including, but not limited to, orthogonally decomposing the prior GP to include extra inducing points
\citep{pmlr-v108-shi20b}, introducing a hierarchical prior over inducing variables to use a massive number of inducing
variables without additional computational cost \citep{pmlr-v139-tran21a}, placing a point process prior over inducing
points that enable the model to learn how many inducing variables to include \citep{pmlr-v161-uhrenholt21a}, etc.
These methods commonly focus on improving the sparse GP prior itself
to include more inducing points for more accurate approximations but are all limited to SVI.
\citet{pmlr-v162-jafrasteh22a} instead introduced amortized variational inference
(AVI, \citet{margossian2024amortized}) to sparse GPs, which learns a mapping from inputs to variational parameters.
This method uses input dependent inducing points that are more informative than independent ones,
thereby drastically reducing the number of inducing variables required and enabling faster inference.

As mentioned above, DGPs are richer models than their shallow counterparts and can hopefully represent more complex
functions, but it suffers from poorer scalability and some pathologies caused by the hierarchical architecture.
Inspired by \citet{pmlr-v162-jafrasteh22a},
we present \emph{amortized variational inference for deep Gaussian processes} in this work\footnote{
	Source code available at \url{https://github.com/qxxmu/DGP-AVI}
}.
Specifically, we leverage the modern formulation of DGPs \citep{NIPS2017_82089746}
and amortize the variational parameters of each layer with the outputs from the previous layer.
We also consider a fully parametric DGP model with deterministic quadrature-like features
\citep{pmlr-v124-jankowiak20a} for stabler amortized variational parameters.
As the original inference function, i.e., a multilayer perceptron (MLP), in \citet{pmlr-v162-jafrasteh22a}
can be problematic for inducing points, we suggest to modify the inference function into independent
affine transformations for inducing points and two MLPs for posterior means and covariances.
We reformulate the evidence lower bound (ELBO) to be optimized and propose three approximation strategies
integrated with different amortization rules.
The resulting method has a non-degenerate input dependent prior with more expressive power than
conventional deep sparse GP priors conditioned on independent inducing points.
Critically, amortization will not degrade posterior approximations in our method,
but instead formulating a novel marginal posterior of high capability at less computational cost.

\section{Background}

In this section we present necessary background on sparse GPs with variational inference (VI)
and demonstrate how VI can be extended to deep Gaussian processes (DGPs) with sparse GP layers.

\subsection{Gaussian process models} \label{Sec_GP}

A Gaussian process (GP, \citet{10.7551/mitpress/3206.001.0001}) represents a distribution denoted as \(\mcal{GP}\)
over a stochastic function \(f(\cdot):\mcal{X}\to\mbb{R}\) defined over an input domain \(\mcal{X}\),
\begin{equation} \label{GP}
	f(\cdot)\sim\mcal{GP}\bigl(m(\cdot),k(\cdot,\cdot')\bigr),
\end{equation}
where \(m(\cdot)=\mbb{E}[f(\cdot)]\) is the mean function, and
\(k(\cdot,\cdot')=\mbb{E}[(f(\cdot)-m(\cdot))(f(\cdot')-m(\cdot'))]\) is the covariance function, a.k.a. kernel.
If we evaluate the GP at an arbitrary finite subset \(\mbf{X}=\{\bsb{x}_n\}_{n=1}^N\) of \(\mcal{X}\),
we would obtain an \(N\)-dimensional Gaussian distribution
\begin{equation} \label{f}
	\mbf{f}\sim\mcal{N}\bigl(\mbf{m}_\mbf{f},\mbf{K}_\mbf{ff}\bigr),
\end{equation}
where \(\mbf{m}_\mbf{f}[n]=m(\bsb{x}_n)\)
and \(\mbf{K}_\mbf{ff}[n,n']=k(\bsb{x}_n,\bsb{x}_{n'})\).

In probabilistic machine learning GPs offer rich nonparametric function priors
with flexible mean and covariance functions that encode prior information about the generative process.
Modelers can adapt GPs to different task scenarios with careful choices of likelihoods.
In this work we consider the prototypical cases of univariate regression and binary classification.
The joint density takes the form
\begin{equation} \label{gpm}
	p(\bsb{y},\mbf{f}|\mbf{X})
	=\underbrace{p(\bsb{y}|\mbf{f})}_\text{likelihood}
	\underbrace{p(\mbf{f}|\mbf{X})}_\text{GP prior},
\end{equation}
where the GP prior is given by Eq. \ref{f}.
Both models share the same likelihoods with their counterparts in generalized linear models.
\paragraph{Univariate regression}
Regression problems have Gaussian likelihoods (usually homoskedastic), i.e.,
an observation \(y_n\) is generated from a Gaussian distribution with mean \(f_n\)
and variance \(\sigma_{obs}^2\) that describes the noise level of observations,
\begin{equation} \label{gauss_ll}
	p(\bsb{y}|\mbf{f})=\mcal{N}\bigl(\bsb{y}|\mbf{f},\sigma_{obs}^2\mbf{I}\bigr).
\end{equation}
\paragraph{Binary classfication}
Binary classification problems have Bernoulli likelihoods, i.e.,
\(y_n\) is assumed to be generated from a Bernoulli distribution,
where an inverse probit link function \enquote{squashes} \(f_n\) to the mean,
\begin{equation}
	\begin{split}
		\Phi(t) & =\int_{-\infty}^{t}\mcal{N}\bigl(a|0,1\bigr)\,da, \\
		p(y_n|f_n) & =\text{Bernoulli}\bigl(y_n|\Phi(f_n)\bigr).
	\end{split}
\end{equation}

The marginal likelihood is obtained by integrating out the latent function values \(\mbf{f}\) in Eq. \ref{gpm},
\begin{equation} \label{marginal_ll}
	p(\bsb{y}|\mbf{X})=\int p(\bsb{y}|\mbf{f})p(\mbf{f}|\mbf{X})\,d\mbf{f}.
\end{equation}
Eq. \ref{marginal_ll} can be computed analytically with Gaussian likelihoods \citep{10.7551/mitpress/3206.001.0001},
and a miscellany of approximation schemes have been proposed for classification problems \citep{JMLR:v9:nickisch08a}.
However, all these computations have a cubic complexity of \(\mcal{O}(N^3)\) w.r.t. sample size \(N\),
dominated by the operation of matrix inversion.

\subsection{Sparse Gaussian processes} \label{Sec_SGP}

Now we introduce a popular approximation to the \emph{exact} GP in the above section.
An additional finite subset \(\mbf{Z}=\{\bsb{z}_m\}_{m=1}^M\) of \(\mcal{X}\) is introduced
s.t. \(\mbf{X}\) and \(\mbf{Z}\) are mutually exclusive.
Similarly to Eq. \ref{f}, evaluating the GP at \(\{\mbf{X},\mbf{Z}\}\) yields a joint Gaussian distribution
\begin{equation} \label{fu}
	\begin{bmatrix}
		\mbf{f} \\ \mbf{u}
	\end{bmatrix}
	\sim\mcal{N}\left(
		\begin{bmatrix}
			\mbf{m}_\mbf{f} \\ \mbf{m}_\mbf{u}
		\end{bmatrix},
		\begin{bmatrix}
			\mbf{K}_\mbf{ff} & \mbf{K}_\mbf{uf} \\
			\mbf{K}_\mbf{fu} & \mbf{K}_\mbf{uu}
		\end{bmatrix}
	\right),
\end{equation}
where \(\mbf{K}_\mbf{fu}[n,m]=\mbf{K}_\mbf{uf}[m,n]=k(\bsb{x}_n,\bsb{z}_m)\).
The conditional distribution of \(\mbf{f}\) conditioned on \(\mbf{u}\) is also Gaussian,
\begin{equation} \label{f|u}
	\mbf{f}|\mbf{u}\sim\mcal{N}\bigl(\tilde{\mbf{m}}_\mbf{f},\tilde{\mbf{K}}_\mbf{ff}\bigr),
\end{equation}
where
\begin{align*}
	\tilde{\mbf{m}}_\mbf{f} & =\mbf{m}_\mbf{f}+\mbf{K}_\mbf{fu}\mbf{K}_\mbf{uu}^{-1}(\mbf{u}-\mbf{m}_\mbf{u}), \\
	\tilde{\mbf{K}}_\mbf{ff} & =\mbf{K}_\mbf{ff}-\mbf{K}_\mbf{fu}\mbf{K}_\mbf{uu}^{-1}\mbf{K}_\mbf{uf}.
\end{align*}
Note that \(\mbf{X}\) can take any location in \(\mcal{X}\) except for the finitely many points \(\mbf{Z}\),
which yields a conditional GP
\begin{equation} \label{SGP}
	f(\cdot)|\mbf{u}\sim\mcal{GP}\bigl(
		m(\cdot)+\mbf{k}_{\cdot\mbf{u}}\mbf{K}_\mbf{uu}^{-1}(\mbf{u}-\mbf{m}_\mbf{u}),
		k(\cdot,\cdot')-\mbf{k}_{\cdot\mbf{u}}\mbf{K}_\mbf{uu}^{-1}\mbf{k}_{\mbf{u}\cdot'}
	\bigr),
\end{equation}
where \(\mbf{k}_{\cdot\mbf{u}}[m]=k(\cdot,\bsb{z}_m)\) and \(\mbf{k}_{\mbf{u}\cdot'}[m]=k(\bsb{z}_m,\cdot')\).

When the evaluation points \(\mbf{Z}\) are treated as parameters, we arrive at the definition of a sparse GP
(SGP, \citet{pmlr-v5-titsias09a}).
In this context, \(\mbf{Z}\) are termed inducing points
and \(\mbf{u}\) the inducing variables that depend on \(\mbf{Z}\).
Sparse GPs improve the computational complexity of GP models with scalable inference algorithms,
as elaborated in the section below.

\subsection{Variational inference for sparse Gaussian processes} \label{Sec_SGP_VI}

With sparse approximations, the GP prior in Eq. \ref{gpm} can be augmented as
\begin{equation} \label{sgp_prior}
	p(\mbf{f}|\mbf{X})\to p(\mbf{f}|\mbf{u},\mbf{X},\mbf{Z})p(\mbf{u}|\mbf{Z})
\end{equation}
As mentioned above, the posterior GP can be inferred via Bayes' rule for an exact GP prior
when the likelihood is Gaussian.
However, a sparse GP does not enjoy closed-form solutions in the general case.
The posterior over latent function values \(\mbf{f}\) and inducing variables \(\mbf{u}\)
is analytically intractable, necessitating approximate techniques during inference.
A convenient approach is variational inference (VI, \citet{doi:10.1080/01621459.2017.1285773}),
with which several algorithms for scalable GP inference has been constructed.

The idea behind VI is to posit a family of densities as approximations to the analytically intractable posterior
and find the best candidate by minimizing the Kullback-Leibler (KL) divergence to the posterior,
so VI actually casts the problem of Bayesian inference as an optimization problem.
\citet{JMLR:v14:hoffman13a} developed stochastic variational inference (SVI),
a stochastic optimization algorithm for factorized (or mean-field) variational inference (FVI)
that allows for optimization on large datasets with stochastic gradient-based methods.

\subsubsection{SVGP} \label{Sec_SVGP}

A sparse variational GP (SVGP, \citet{10.5555/3023638.3023667}) is obtained by applying SVI to sparse GPs.
The primary idea is to approximate the posterior with another density \(q(\mbf{f},\mbf{u})\).
\citet{10.5555/3023638.3023667} followed the formulation of variational free energy
(VFE, \citet{pmlr-v5-titsias09a, NIPS2016_7250eb93}) approximation,
i.e., to keep the conditioning process \(p(\mbf{f}|\mbf{u},\mbf{X},\mbf{Z})\)
in joint GP prior intact and introduce an extra Gaussian density
\(q(\mbf{u})=\mcal{N}(\mbf{u}|\bsb{\mu},\bsb{\Sigma})\)
to approximate the posterior over inducing variables, then the joint variational posterior is
\begin{equation}
	q(\mbf{f},\mbf{u}|\mbf{X},\mbf{Z})=p(\mbf{f}|\mbf{u},\mbf{X},\mbf{Z})q(\mbf{u}).
\end{equation}
The evidence lower bound (ELBO), i.e., the objective function to be optimized,
can be formulated by definition as
\begin{equation} \label{raw_elbo}
	\begin{split}
		\mcal{L}_\text{svgp} & =\mbb{E}_{q(\mbf{f},\mbf{u}|\mbf{X},\mbf{Z})}\left[
			\ln\frac{p(\bsb{y}|\mbf{f}){\color{gray}p(\mbf{f}|\mbf{u},\mbf{X},\mbf{Z})}p(\mbf{u}|\mbf{Z})}
			{{\color{gray}p(\mbf{f}|\mbf{u},\mbf{X},\mbf{Z})}q(\mbf{u})}
		\right] \\
		& =\mbb{E}_{q(\mbf{f},\mbf{u}|\mbf{X},\mbf{Z})}\bigl[\ln p(\bsb{y}|\mbf{f})\bigr]
		-D_\text{KL}\bigl(q(\mbf{u})\|p(\mbf{u}|\mbf{Z})\bigr)
		\leq\ln p(\bsb{y}|\mbf{X}).
	\end{split}
\end{equation}
Be aware that the targets \(\bsb{y}\) depend solely on the latent function values \(\mbf{f}\),
then the inducing variables \(\mbf{u}\) can be integrated out in the variational posterior,
yielding a marginal Gaussian density
\begin{equation} \label{sgp_marginal_posterior}
	q(\mbf{f}|\mbf{X},\mbf{Z})=\int p(\mbf{f}|\mbf{u},\mbf{X},\mbf{Z})q(\mbf{u})\,d\mbf{u}
	=\mcal{N}\bigl(\mbf{f}|\ddot{\mbf{m}}_\mbf{f},\ddot{\mbf{K}}_\mbf{ff}\bigr),
\end{equation}
where
\begin{align*}
	\ddot{\mbf{m}}_\mbf{f} & =\mbf{m}_\mbf{f}+\mbf{K}_\mbf{fu}\mbf{K}_\mbf{uu}^{-1}
	(\bsb{\mu}-\mbf{m}_\mbf{u}), \\
	\ddot{\mbf{K}}_\mbf{ff} & =\tilde{\mbf{K}}_\mbf{ff}+\mbf{K}_\mbf{fu}\mbf{K}_\mbf{uu}^{-1}
	\bsb{\Sigma}\mbf{K}_\mbf{uu}^{-1}\mbf{K}_\mbf{uf}.
\end{align*}
For Gaussian likelihoods, the expectation term in Eq. \ref{raw_elbo} has analytic solution
\begin{equation}
	\begin{split}
		\mbb{E}_{q(\mbf{f},\mbf{u}|\mbf{X},\mbf{Z})}\bigl[\ln p(\bsb{y}|\mbf{f})\bigr]
		& =\mbb{E}_{q(\mbf{f}|\mbf{X},\mbf{Z})}\Bigl[
			\ln \mcal{N}\bigl(\bsb{y}|\mbf{f},\sigma_{obs}^2\mbf{I}\bigr)
		\Bigr]=\mcal{N}\bigl(\bsb{y}|\ddot{\mbf{m}}_\mbf{f},\sigma_{obs}^2\mbf{I}\bigr)
		-\frac{\operatorname{Tr}\ddot{\mbf{K}}_\mbf{ff}}{2\sigma_{obs}^2}.
	\end{split}
\end{equation}
Generally, non-Gaussian likelihoods have no analytic solutions (e.g., Bernoulli likelihoods).
Nevertheless, expectations w.r.t. a Gaussian density can easily be approximated by Gauss-Hermite quadrature,
as suggested by \citet{pmlr-v38-hensman15}.
Note that the objective in Eq. \ref{raw_elbo} factorizes across datapoints,
which enables data subsampling during optimization.
As Eq. \ref{sgp_marginal_posterior} shows, the computation of a sparse GP is dominated by the precision matrix
\(\mbf{K}_\mbf{uu}^{-1}\). The complexity scales down to
\(\mcal{O}(M^3)\) where \(M\ll N\) introduces \emph{sparsity} to a GP model.

We would elucidate that besides the approximate mean \(\bsb{\mu}\) and covariance \(\bsb{\Sigma}\),
the inducing points \(\mbf{Z}\) are also variational parameters in this context.
This is a bit subtle because a sparse GP itself is an approximation to the corresponding exact GP.\footnote{
	VI for sparse GPs differs from ordinary formulations in that
	the prior and the variational posterior share the conditioning process,
	and we refer readers to \citet{leibfried2022tutorial} for additional interpretations.
}

\subsection{Deep Gaussian processes} \label{Sec_DGP}

Deep Gaussian processes (DGPs, \citet{pmlr-v31-damianou13a})
are deep hierarchies obtained by stacking a sequence of GPs
in which the outputs of a GP layer become the inputs of the subsequent GP layer.
DGPs are flexible models with promisingly greater expressivity than their single-layer counterparts,
despite the intractability in inference.

\subsubsection{Doubly-stochastic variational inference}

\citet{NIPS2017_82089746} defined a modern formulation of DGP by introducing inducing variables to the GP layers,
resulting in a compositional function prior
\begin{equation}
	{\prod}_{l=1}^{L}p(\mbf{F}^l|\mbf{U}^l,\mbf{F}^{l-1},\mbf{Z}^{l-1})p(\mbf{U}^l|\mbf{Z}^{l-1}),
\end{equation}
where \(\mbf{F}^l=\{\mbf{f}_n^l\}_{n=1}^N\), and \(\mbf{F}^0:=\mbf{X}\).
The inducing variables are \(\mbf{U}^l=\{\mbf{u}_d^l\}_{d=1}^{D^l}\),
where \(D^l\) denotes the output dimension at the \(l\)-th layer, and \(D^0\) is the input dimension.
The GP prior for each layer is defined analogously to the RHS of Eq. \ref{sgp_prior}.
When Gaussian densities \(\{q(\mbf{U}^l)=\mcal{N}(\mbf{U}^l|\bsb{\mu}^l,\bsb{\Sigma}^l)\}_{l=1}^L\)
are introduced to approximate the posteriors over the inducing variables,
the variational posterior takes the form
\begin{equation} \label{dgp_posterior}
	q\bigl(\{\mbf{F}^l\}_{l=1}^L,\{\mbf{U}^l\}_{l=1}^L|\mbf{X},\{\mbf{Z}^{l-1}\}_{l=1}^L\bigr)
	={\prod}_{l=1}^{L}p(\mbf{F}^l|\mbf{U}^l,\mbf{F}^{l-1},\mbf{Z}^{l-1})q(\mbf{U}^l).
\end{equation}
Recall that the likelihood only depends on the outputs of the final layer,
then the marginalizing Eq. \ref{dgp_posterior} w.r.t. \(\mbf{F}^L\) yields
\begin{equation} \label{dgp_marginal_posterior}
	\begin{split}
		q\bigl(\mbf{F}^L|\mbf{X},\{\mbf{Z}^{l-1}\}_{l=1}^L\bigr)
		& =\int\Bigl[{\prod}_{l=1}^Lp(\mbf{F}^l|\mbf{U}^l,\mbf{F}^{l-1},\mbf{Z}^{l-1})q(\mbf{U}^l)\Bigr]
		\Bigl[{\prod}_{l=1}^Ld\mbf{U}^l\Bigr]\Bigl[{\prod}_{l=1}^{L-1}d\mbf{F}^l\Bigr] \\
		& =\int\Bigl[{\prod}_{l=1}^Lq(\mbf{F}^l|\mbf{F}^{l-1},\mbf{Z}^{l-1})\Bigr]
		\Bigl[{\prod}_{l=1}^{L-1}d\mbf{F}^l\Bigr], \\
	\end{split}
\end{equation}
where each \(q(\mbf{F}^l|\mbf{F}^{l-1},\mbf{Z}^{l-1})\)
is a Gaussian density in analogy to Eq. \ref{sgp_marginal_posterior}.
The integral in Eq. \ref{dgp_marginal_posterior} is analytically intractable
but straightforward to approximate with Monte Carlo sampling, which results in a finite Gaussian mixture.

The \emph{doubly stochastic} (DS)-ELBO is given by
\begin{equation}
	\mcal{L}_\text{dgp}=\mbb{E}_{q(\mbf{F}^L|\mbf{X},\{\mbf{Z}^{l-1}\}_{l=1}^L)}
	\bigl[\ln p(\mbf{Y}|\mbf{F}^L)\bigr]
	-{\sum}_{l=1}^{L}D_\text{KL}\bigl(q(\mbf{U}^l)\|p(\mbf{U}^l|\mbf{Z}^{l-1})\bigr),
\end{equation}
which is also factorizable across datapoints.
Aside from the mini-batching of input data as in SVI, random sampling to approximate the marginal posterior
also brings stochasticity to the model.

\section{Amortized variational inference for sparse Gaussian processes} \label{Sec_SGP_AVI}

The key innovation of \citet{pmlr-v162-jafrasteh22a} is to amortize the computation of
the variational parameters with neural networks, which produces input dependent inducing variables
\(\{\mbf{u}_n\}_{n=1}^N\) of a smallish number \(M\).
Here we review this method in terms of amortized variational inference (AVI, \citet{margossian2024amortized}).

\begin{figure} [ht]
    \centering
	\begin{tikzpicture}
		[
			node distance=5mm and 5mm,
			round/.style={circle, draw=black, minimum size=8mm},
			plain/.style={},
			obs/.style={circle, draw=black, fill=lightgray, minimum size=8mm},
		]
		\begin{scope}
			[node distance=8mm and 8mm]
			\node[round](f){\(f_n\)};
			\node[obs](x)[left=of f]{\(\bsb{x}_n\)};
			\node[obs](y)[below=of f]{\(y_n\)};
			\node[round](u)[above=of f]{\(\mbf{u}\)};
		\end{scope}
		\node[plain](Z)[left=of u, yshift=-5mm]{\(\mbf{Z}\)};
		\node[plain](mean)[left=of u]{\(\bsb{\mu}\)};
		\node[plain](cov)[left=of u, yshift=5mm]{\(\bsb{\Sigma}\)};
		\node[plain](theta)[below right=of u]{\(\theta\)};
		\node[plain](sigma)[above right=of y]{\(\sigma_{obs}^2\)};
		\plate[plain]{plate1} {(x)(f)(y)} {\(N\)}
		\draw[-Stealth](x)--(f);
		\draw[-Stealth](f)--(y);
		\draw[-Stealth](u)--(f);
		\draw[-Stealth](Z)--(u);
		\draw[-Stealth](Z)--(f);
		\draw[-Stealth](mean)--(u);
		\draw[-Stealth](cov)--(u);
		\draw[-Stealth](theta)--(u);
		\draw[-Stealth](theta)--(f);
		\draw[-Stealth](sigma)--(y);
	\end{tikzpicture}
	\hspace{5mm}
	\begin{tikzpicture}
		[
			node distance=5mm and 5mm,
			round/.style={circle, draw=black, inner sep=0.2mm, minimum size=8mm},
			plain/.style={},
			obs/.style={circle, draw=black, fill=lightgray, inner sep=0.2mm, minimum size=8mm},
		]
		\begin{scope}
			[node distance=8mm and 8mm]
			\node[round](f){\(f_n\)};
			\node[obs](x)[left=of f]{\(\bsb{x}_n\)};
			\node[obs](y)[below=of f, xshift=8mm]{\(y_n\)};
			\node[round](u)[above=of f, xshift=8mm]{\(\mbf{u}_n\)};
		\end{scope}
		\node[plain](Z)[left=of u, yshift=-5mm]{\(\mbf{Z}_n\)};
		\node[plain](mean)[left=of u]{\(\bsb{\mu}_n\)};
		\node[plain](cov)[left=of u, yshift=5mm]{\(\bsb{\Sigma}_n\)};
		\node[plain](theta)[below right=of u]{\(\theta\)};
		\node[plain](sigma)[above right=of y]{\(\sigma_{obs}^2\)};
		\plate[plain]{plate1} {(x)(mean)(cov)(u)(f)(y)} {\(N\)}
		\draw[-Stealth](x)--(f);
		\draw[-Stealth](f)--(y);
		\draw[-Stealth](u)--(f);
		\draw[-Stealth](Z)--(u);
		\draw[-Stealth](Z)--(f);
		\draw[-Stealth](mean)--(u);
		\draw[-Stealth](cov)--(u);
		\draw[-Stealth](theta)--(u);
		\draw[-Stealth](theta)--(f);
		\draw[-Stealth](sigma)--(y);
		\draw[ForestGreen, -Stealth](x.north)..controls+(up:5mm) and +(left:5mm)..(Z.west);
		\draw[ForestGreen, -Stealth](x.north)..controls+(up:6mm) and +(left:6mm)..(mean.west);
		\draw[ForestGreen, -Stealth](x.north)..controls+(up:7mm) and +(left:7mm)..(cov.west);
	\end{tikzpicture}
	\caption{
		Probabilistic graphical models of \textit{left}: SVGP and \textit{right}: IDSGP.
		SVGP uses independent inducing variables \(\mbf{u}\) for approximation.
		In IDSGP an inference function denoted by the green arrows maps each input \(\bsb{x}_n\) to
		the input dependent inducing variables \(\mbf{u}_n\).
	}
	\label{fig1}
\end{figure}
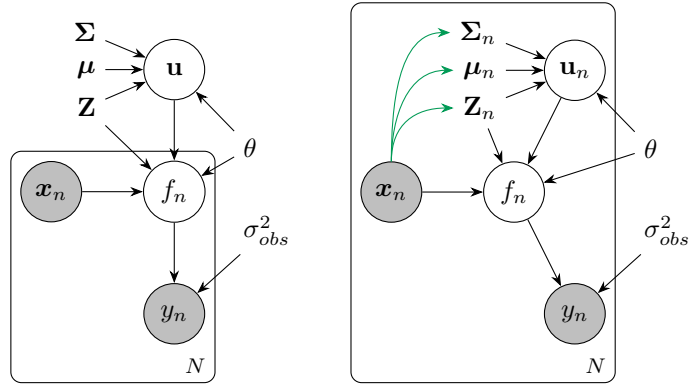

Traditional VI techniques directly fit parametric variational distributions for latent variables.
Conversely, AVI learns a common inference function which maps each observation
to its corresponding latent variable's approximate posterior.
A sparse GP prior is conditioned on the inducing variables as in Eq. \ref{sgp_prior},
where \(\mbf{u}\) are global latent variables to all observation pairs \(\{\bsb{x}_n,y_n\}_{n=1}^N\).
Therefore, AVI is inapplicable to the original formulation of sparse GPs.
The modified structure of sparse GPs implied by \citet{pmlr-v162-jafrasteh22a} uses \emph{local}
inducing variables \(\mbf{u}_n\) for each input (Fig. \ref{fig1}) and thereby enables amortization.
One can verify that such formulation is a simple hierarchical model as defined in \citet{margossian2024amortized},
and there thus exists an ideal inference function that maps each observation to the variational parameters.\footnote{
	See \citet{NEURIPS2021_b28d7c6b, margossian2024amortized} for proof.
}

\citet{pmlr-v162-jafrasteh22a} proposed to employ a neural network as inference function,
\begin{equation} \label{inference_func}
	\{\mbf{Z}_n,\bsb{\mu}_n,\mbf{L}_n\}=g_\phi(\bsb{x}_n),
\end{equation}
where \(g_\phi\) is the neural network parameterized by \(\phi\),
and \(\mbf{L}_n\) is the Cholesky factor of covariance matrix \(\bsb{\Sigma}_n=\mbf{L}_n\mbf{L}_n^\msf{T}\).
The domain of the inference function is the input domain \(\mcal{X}\),
then all computations are \emph{input dependent} as Fig. \ref{fig1} shows.

IDSGP improves the computational cost of sparse GPs with much fewer inducing variables.
Intuitively, with an ideal inference function we produce a small number of optimal inducing points for each
input datapoint, which is more efficient than a large number of global inducing points for all inputs.
Another benefit that AVI brings is faster convergence to the optimal solution.
Moreover, IDSGP performs better than sparse GPs owing to the neural network of high capability and flexibility.

\subsection{Affine transformations as inference functions} \label{Sec_affine}

Sparse GPs with SVI generalize well in held-out data. IDSGP, however, may suffer from overfitting in some problems.
This tendency to overfit is attributable to two reasons:
Firstly, the modified formulation uses separate inducing points rather than global ones for different inputs,
then the loss of generalizability is inevitable.
Secondly, the limitations of AVI per se worsen the problem.
As argued in \citet{10.1613/jair.1.14258}, an insufficient inference function cannot close
the amortization gap \citep{pmlr-v80-cremer18a}, while an over-expressive one increases
the generalization gap \citep{10.5555/3600270.3602212}.

We also observe that IDSGP can show pathological behaviour as the depth of
the amortization neural network increases (see Appendix \ref{App_infer_func}).
A deep amortization neural network tends to map the input domain \(\mcal{X}\) to a subspace of itself
and ignore the relative position of each input.
The consequent degenerate inducing points can severely reduce the expressivity of the sparse GP prior.
We blame this on the complex nonlinearities in very deep neural networks,
which result in highly non-injective functions \citep{pmlr-v33-duvenaud14}.
On the contrary, a shallow neural network is inadequate to represent the real function over
the parameters \(\{\bsb{\mu}_n,\mbf{L}_n\}\) of \(q(\mbf{u}_n)\).
To encourage the inference function to produce non-degenerate inducing points
while retaining the accuracy of posterior approximation, we propose to reformulate it as follows:
\begin{equation} \label{affine_func}
	\mbf{Z}_n=\bm{\mcal{A}}(\bsb{x}_n),
	\quad\bsb{\mu}_n=g_\phi(\bsb{x}_n),
	\quad\mbf{L}_n=h_\psi(\bsb{x}_n),
\end{equation}
where \(g\) and \(h\) are two neural networks parameterized by \(\phi\) and \(\psi\) respectively.
The inducing points are mapped by \(M\) affine transformations
\(\bm{\mcal{A}}(\cdot)=[\mcal{A}_1(\cdot),\ldots,\mcal{A}_M(\cdot)]^\msf{T}\),
with each \(\mcal{A}_m\) given by
\begin{equation}
	\mcal{A}_m(\bsb{x}_n)=\bsb{z}_{n,m}=\mbf{W}_m\bsb{x}_n+\mbf{b}_m.
\end{equation}
where the invertible transformation matrix \(\mbf{W}_m\) and vector \(\mbf{b}_m\in\mcal{X}\) are learnable parameters,
then \(\bsb{z}_{n,m}\) can take any location in \(\mcal{X}\).
We find such formulation superior to the single neural network \citet{pmlr-v162-jafrasteh22a}
used as in Eq. \ref{inference_func} (see Appendix \ref{App_infer_func} for details).

\section{Amortized variational inference for deep Gaussian processes} \label{Sec_DGP_AVI}

The compositional structure of GP priors allows the DGP posterior to represent a larger class of distributions
beyond the Gaussian posterior of shallow GPs, e.g.,
the doubly stochastic DGP (DS-DGP, \citet{NIPS2017_82089746}) defines the variational posterior
to be a continuous Gaussian mixture which is in practice approximated by a finite Gaussian mixture.
In this section we describe how AVI can be applied to DGPs to obtain richer models
than shallow sparse GPs with AVI.

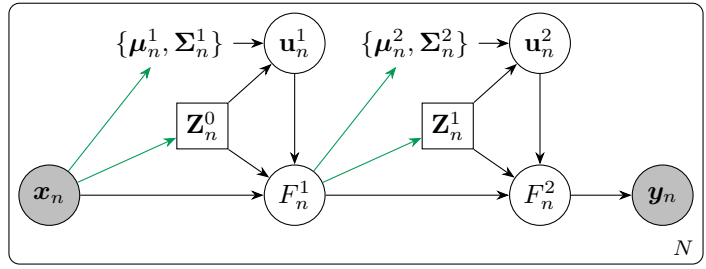
\begin{figure} [ht]
	\centering
	\begin{tikzpicture}
		[
			node distance=8mm and 8mm,
			round/.style={circle, draw=black, inner sep=0.2mm, minimum size=8mm},
			plain/.style={},
			obs/.style={circle, draw=black, fill=lightgray, inner sep=0.2mm, minimum size=8mm},
		]
		\node[round](f1){\(F_n^1\)};
		\node[round, draw=none](i0)[left=of f1]{};
		\node[round, draw=none](i1)[right=of f1]{};
		\node[round](f2)[right=of i1]{\(F_n^2\)};
		\node[obs](x)[left=of i0]{\(\bsb{x}_n\)};
		\node[obs](y)[right=of f2]{\(\bsb{y}_n\)};
		\node[round](u1)[above=of f1, yshift=4mm]{\(\mbf{u}_n^1\)};
		\node[round](u2)[above=of f2, yshift=4mm]{\(\mbf{u}_n^2\)};
		\node[rectangle, draw](z0)[above=of i0, xshift=4mm, yshift=-6mm]{\(\mbf{Z}_n^0\)};
		\node[](ms1)[left=of u1, xshift=4mm]{\(\{\bsb{\mu}_n^1,\bsb{\Sigma}_n^1\}\)};
		\node[rectangle, draw](z1)[above=of i1, xshift=4mm, yshift=-6mm]{\(\mbf{Z}_n^1\)};
		\node[](ms2)[left=of u2, xshift=4mm]{\(\{\bsb{\mu}_n^2,\bsb{\Sigma}_n^2\}\)};
		\plate[plain]{plate1} {(x)(u1)(y)} {\(N\)}
		\draw[-Stealth](x)--(f1);
		\draw[-Stealth](f1)--(f2);
		\draw[-Stealth](f2)--(y);
		\draw[-Stealth](u1)--(f1);
		\draw[-Stealth](u2)--(f2);
		\draw[ForestGreen, -Stealth](x)--(z0);
		\draw[-Stealth](z0.north east)--(u1);
		\draw[ForestGreen, -Stealth](x)--(ms1);
		\draw[-Stealth](ms1)--(u1);
		\draw[-Stealth](z0.south east)--(f1);
		\draw[ForestGreen, -Stealth](f1)--(z1);
		\draw[-Stealth](z1.north east)--(u2);
		\draw[-Stealth](z1.south east)--(f2);
		\draw[ForestGreen, -Stealth](f1)--(ms2);
		\draw[-Stealth](ms2)--(u2);
	\end{tikzpicture}
	\caption{
		Probabilistic graphical model of 2-layered DGP with AVI.
		Inference functions denoted by the green arrows map the outputs \(F_n^{l-1}\) at the \((l-1)\)-th layer
		to the variational parameters \(\{\mbf{Z}_n^l,\bsb{\mu}_n^l,\bsb{\Sigma}_n^l\}\) of the \(l\)-th layer.
	}
	\label{fig2}
\end{figure}

We follow DS-DGP to construct a composition of layer-wise variational posteriors
and amortize the computation of variational parameters.
The basic idea is to map the outputs of a GP layer to the variational parameters of the subsequent layer,
then the conditional variational posterior of each layer is similar to that of a shallow sparse GP with AVI.
However, amortization is more challenging in DGPs since the outputs of intermediate layers are
(Gaussian) random variables instead of deterministic values.
Let \(\mcal{P}(\cdot)\) denote an operator on latent function values \(\{F_n^{l-1}\}_{n=1}^N\),
then the \(l\)-th layer has amortized variational parameters
\begin{equation} \label{infer_func}
	\begin{split}
		& \mbf{Z}_n^{l-1}=\bm{\mcal{A}}^l(\mcal{P}(F_n^{l-1})), \\
		& \bsb{\mu}_n^l=\{\mu_{d,n}^l\}_{d=1}^{D^l}=g_{\phi_l}(\mcal{P}(F_n^{l-1})), \\
		& \mbf{L}_n^l=\{L_{d,n}^l\}_{d=1}^{D^l}=h_{\psi_l}(\mcal{P}(F_n^{l-1})),
	\end{split}
\end{equation}
where \(\bm{\mcal{A}}^l\), \(g_{\phi_l}\), \(h_{\psi_l}\) are defined analogously to
those in Eq. \ref{affine_func}, and \(L_{d,n}^l\) is the Cholesky factor of
\(\Sigma_{d,n}^l=L_{d,n}^l[L_{d,n}^l]^\msf{T}\).
In this way, the computations for each layer become completely dependent on the outputs of the
previous layer with no additional separate variational factors involved, as Fig. \ref{fig2} shows.

\subsection{Input dependent prior and variational posterior}

The model prior is compositional as in \citet{NIPS2017_82089746},
then for a single datapoint \(F_n^0:=\bsb{x}_n\) we write the sparse GP prior \(f^{(l)}(\cdot)\) over
the corresponding output point \(F_n^l\) at the \(l\)-th layer as follows,
\begin{align}
	p(\mbf{U}_n^l|F_n^{l-1},\mbf{Z}_n^{l-1})
	& ={\prod}_{d=1}^{D^l}p(\mbf{u}_{d,n}^l|F_n^{l-1},\mbf{Z}_n^{l-1})={\prod}_{d=1}^{D^l}
	\mcal{N}\bigl(\mbf{u}_{d,n}^l|m_d^l(\mbf{Z}_n^{l-1}),\mbf{K}_{\mbf{uu}_{d,n}}^l\bigr), \\
	p(F_n^l|\mbf{U}_n^l,F_n^{l-1},\mbf{Z}_n^{l-1})
	& ={\prod}_{d=1}^{D^l}p(f_{d,n}^l|\mbf{u}_{d,n}^l,F_n^{l-1},\mbf{Z}_n^{l-1})
	={\prod}_{d=1}^{D^l}\mcal{N}\bigl(f_{d,n}^l|\tilde{m}_{f_{d,n}}^l,\tilde{K}_{f_{d,n}}^l\bigr),
\end{align}
where
\begin{align*}
	\tilde{m}_{f_{d,n}}^l & =m_d^l(F_n^{l-1})+\mbf{K}_{\mbf{fu}_{d,n}}^l
	[\mbf{K}_{\mbf{uu}_{d,n}}^l]^{-1}\bigl(\mbf{u}_{d,n}^l-m_d^l(\mbf{Z}_n^{l-1})\bigr), \\
	\tilde{K}_{f_{d,n}}^l & =k_d^l(F_n^{l-1})
	-\mbf{K}_{\mbf{fu}_{d,n}}^l[\mbf{K}_{\mbf{uu}_{d,n}}^l]^{-1}\mbf{K}_{\mbf{uf}_{d,n}}^l.
\end{align*}
\(m_d^l(\cdot)\) and \(k_d^l(\cdot)\) denote the mean and covariance function
for the \(d\)-th dimension of outputs at the \(l\)-th layer, respectively.
The (cross)-covariances are given by \(\mbf{K}_{\mbf{uu}_{d,n}}^l=k_d^l(\mbf{Z}_n^{l-1})\) and
\(\mbf{K}_{\mbf{uf}_{d,n}}^l=[\mbf{K}_{\mbf{fu}_{d,n}}^l]^\msf{T}=k_d^l(\mbf{Z}_n^{l-1},F_n^{l-1})\).

\begin{figure} [ht]
	\includegraphics[width=15cm, center]{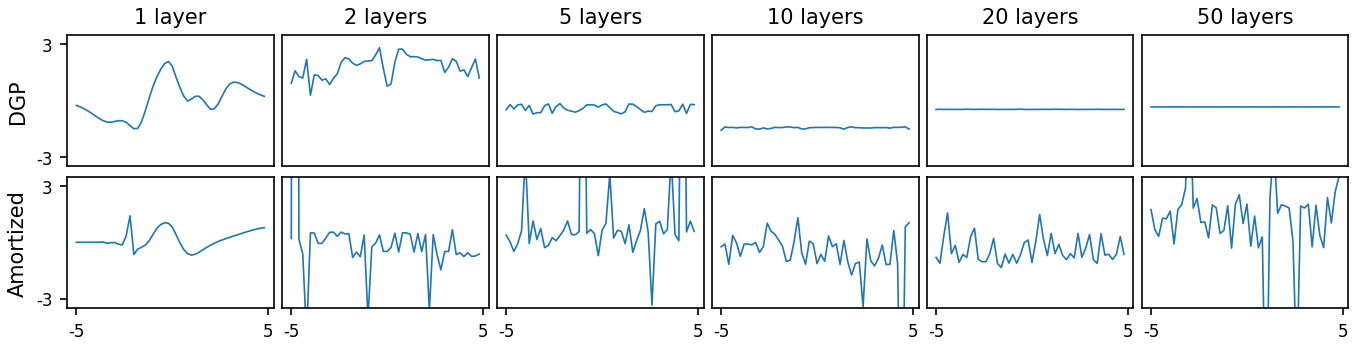}
	\caption{
		Samples \(f^{(1:L)}(\bsb{x})\) successively drawn from 1-d \textit{top}: conventional DGP priors (\(M=128\))
		and \textit{bottom}: amortized DGP priors (\(M=4\)), both with zero mean functions.
		Each column represents the number of layers, where 1 layer corresponds to shallow GPs.
		The function begins to concentrate at some values after a few layers,
		i.e., the prior fails to model functions of interest.
		This degeneration does not occur with amortized DGP priors.
	}
	\label{prior_sample}
\end{figure}

A compositional DGP prior with zero mean functions is pathological as the derivate of the process becomes
very small almost everywhere with rare but very large jumps when the depth increases \citep{pmlr-v33-duvenaud14}.
The function values quickly become either nearly flat or quickly-varying everywhere as Fig. \ref{prior_sample} shows.
In an amortized DGP prior, the mean of each layer is conditioned on input dependent inducing variables,
and thus expressive enough to represent complex mappings with zero mean functions.
Notably, the modified prior does not degenerate as the number of layers increases.

Now we derive the marginal posterior for each GP layer.
The variational posterior over inducing variables is
\begin{equation}
	q(\mbf{U}_n^l|F_n^{l-1})={\prod}_{d=1}^{D^l}q(\mbf{u}_{d,n}^l|F_n^{l-1})
	={\prod}_{d=1}^{D^l}\mcal{N}\bigl(\mbf{u}_{d,n}^l|\mu_{d,n}^l,\Sigma_{d,n}^l\bigr),
\end{equation}
then the posterior over \(F_n^l\) is obtained by marginalizing out the inducing variables analytically
as in Eq. \ref{sgp_marginal_posterior},
\begin{equation}
	\begin{split}
		q(F_n^l|F_n^{l-1},\mbf{Z}_n^{l-1})
		& ={\prod}_{d=1}^{D^l}q(f_{d,n}^l|F_n^{l-1},\mbf{Z}_n^{l-1}) \\
		& ={\prod}_{d=1}^{D^l}\int p(\mbf{u}_{d,n}^l|F_n^{l-1},\mbf{Z}_n^{l-1})
		q(\mbf{u}_{d,n}^l|F_n^{l-1})\,d\mbf{u}_{d,n}^l \\
		& ={\prod}_{d=1}^{D^l}\mcal{N}\bigl(f_{d,n}^l|\ddot{m}_{f_{d,n}}^l,\ddot{K}_{f_{d,n}}^l\bigr),
	\end{split}
\end{equation}
where
\begin{align*}
	\ddot{m}_{f_{d,n}}^l & =m_d^l(F_n^{l-1})+\mbf{K}_{\mbf{fu}_{d,n}}^l
	[\mbf{K}_{\mbf{uu}_{d,n}}^l]^{-1}\bigl(\mu_{d,n}^l-m_d^l(\mbf{Z}_n^{l-1})\bigr), \\
	\ddot{K}_{f_{d,n}}^l & =\tilde{K}_{f_{d,n}}^l
	+\mbf{K}_{\mbf{fu}_{d,n}}^l[\mbf{K}_{\mbf{uu}_{d,n}}^l]^{-1}
	\Sigma_{d,n}^l[\mbf{K}_{\mbf{uu}_{d,n}}^l]^{-1}\mbf{K}_{\mbf{uf}_{d,n}}^l.
\end{align*}
Note that the inducing points \(\mbf{Z}_n^{l-1}\) depend on inputs \(F_n^{l-1}\),
thus notationally redundant to the posterior which can be reduced to \(q(F_n^l|F_n^{l-1})\).
We keep \(\mbf{Z}_n^{l-1}\) in above equations to help emphasize the dependences among variables.

\subsection{Lower bound on the marginal likelihood}

Following \citet{pmlr-v162-jafrasteh22a} and \citet{NIPS2017_82089746}, the ELBO can be formulated as
(see Appendix \ref{derive_dselbo} for a detailed derivation)
\begin{equation} \label{dgp_avi_elbo}
	\begin{split}
		\mcal{L}={\sum}_{n=1}^N\biggl\{ & \mbb{E}_{q(F_n^L|\bsb{x}_n)}\bigl[\ln p(\bsb{y}_n|F_n^L)\bigr] \\
		& -\frac{1}{N}{\sum}_{l=1}^L\mbb{E}_{q(F_n^{l-1}|\bsb{x}_n)}\Bigl[
			D_\text{KL}\bigl(q(\mbf{U}_n^l|F_n^{l-1})\|p(\mbf{U}_n^l|F_n^{l-1})\bigr)
		\Bigr]\biggr\},
	\end{split}
\end{equation}
where \(q(F_n^l|\bsb{x}_n)\) is the marginal posterior over the outputs at the \(l\)-th layers.
The expected log-likelihood terms can be approximated by Monte Carlo integration as in
Eq. \ref{dgp_marginal_posterior},
\begin{equation} \label{ell_approx_1}
	\mbb{E}_{q(F_n^L|\bsb{x}_n)}\bigl[\ln p(\bsb{y}_n|F_n^L)\bigr]\approx\frac{1}{S}
	{\sum}_{s=1}^{S}\mbb{E}_{q(F_{n,s}^L|F_{n,s}^{L-1})}\bigl[\ln p(\bsb{y}_n|F_{n,s}^L)\bigr],
\end{equation}
where we randomly draw \(S\) groups of samples \(\{F_{n,s}^{l-1}\}_{l=1}^L\)
from the compositional variational posterior as \(\prod_{l=1}^{L}q(F_{n,s}^l|F_{n,s}^{l-1})\).
The KL terms --- in contrast to those in conventional methods with independent inducing variables ---
are wrapped in expectations because the inducing variables \(\mbf{U}_n^l\) depend on
\(F_{n,s}^{l-1}\) for \(l\geq 2\).
Analytic solutions are generally unavailable, but hopefully we may approximate them with
Monte Carlo sampling for different choices of operators \(\mcal{P}(\cdot)\) as discussed in the next section. 

\subsection{Amortization strategies under Monte Carlo approximation} \label{Sec_AR}

Note that the variational parameters in the first layer are mapped from deterministic inputs,
which can be viewed as Gaussian random variables with means \(\bsb{x}_n\)'s and zero variances.
So the behaviour of operator \(\mcal{P}\) should be consistent with the first layer.
Here we describe two amortization rules with corresponding approximations to the ELBO.

\begin{figure} [ht]
	\centering
	\begin{tikzpicture}
		[
			node distance=4mm and 4mm,
			round/.style={circle, draw=black, inner sep=0.2mm, minimum size=8mm},
			plain/.style={},
			obs/.style={circle, draw=black, fill=lightgray, inner sep=0.2mm, minimum size=8mm},
		]
		\node[round](u1){\(\mbf{U}^1\)};
		\node[obs](x1)[below=of u1, xshift=-12mm]{\(\bsb{x}_1\)};
		\node[obs](x2)[below=of u1, xshift=12mm]{\(\bsb{x}_2\)};
		\node[round](f111)[above=of u1, xshift=-18mm]{\(F_{11}^1\)};
		\node[round](f112)[above=of u1, xshift=-6mm]{\(F_{12}^1\)};
		\node[round](f121)[above=of u1, xshift=6mm]{\(F_{21}^1\)};
		\node[round](f122)[above=of u1, xshift=18mm]{\(F_{22}^1\)};
		\node[round, draw=none](f1i)[above=of u1]{\(\phantom{F_{n,s}^1}\)};
		\node[round](u2)[above=of f1i]{\(\mbf{U}^2\)};
		\node[round](f211)[above=of u2, xshift=-18mm]{\(F_{11}^2\)};
		\node[round](f212)[above=of u2, xshift=-6mm]{\(F_{12}^2\)};
		\node[round](f221)[above=of u2, xshift=6mm]{\(F_{21}^2\)};
		\node[round](f222)[above=of u2, xshift=18mm]{\(F_{22}^2\)};
		\node[round, draw=none](f2i)[above=of u2]{\(\phantom{F_{n,s}^2}\)};
		\node[round](f21)[above=of f2i, xshift=-12mm]{\(F_{1}^2\)};
		\node[round](f22)[above=of f2i, xshift=12mm]{\(F_{2}^2\)};
		\node[obs](y1)[above=of f21]{\(\bsb{y}_1\)};
		\node[obs](y2)[above=of f22]{\(\bsb{y}_2\)};
		\draw[-Stealth](x1)--(f111);
		\draw[-Stealth](x1)--(f112);
		\draw[-Stealth](x2)--(f121);
		\draw[-Stealth](x2)--(f122);
		\draw[-Stealth](u1)--(f111);
		\draw[-Stealth](u1)--(f112);
		\draw[-Stealth](u1)--(f121);
		\draw[-Stealth](u1)--(f122);
		\draw[-Stealth](f111)--(f211);
		\draw[-Stealth](f112.120)--(f212.240);
		\draw[-Stealth](f121.60)--(f221.300);
		\draw[-Stealth](f122)--(f222);
		\draw[-Stealth](u2)--(f211);
		\draw[-Stealth](u2)--(f212);
		\draw[-Stealth](u2)--(f221);
		\draw[-Stealth](u2)--(f222);
		\draw[-Stealth](f211)--(f21);
		\draw[-Stealth](f212)--(f21);
		\draw[-Stealth](f221)--(f22);
		\draw[-Stealth](f222)--(f22);
		\draw[-Stealth](f21)--(y1);
		\draw[-Stealth](f22)--(y2);
	\end{tikzpicture}
	\hspace{4mm}
	\vrule
	\hspace{4mm}
	\begin{tikzpicture}
		[
			node distance=4mm and 4mm,
			round/.style={circle, draw=black, inner sep=0.2mm, minimum size=8mm},
			plain/.style={},
			obs/.style={circle, draw=black, fill=lightgray, inner sep=0.2mm, minimum size=8mm},
		]
		\node[round, draw=none](u1){\(\phantom{\mbf{U}^1}\)};
		\node[obs](x1)[below=of u1, xshift=-12mm]{\(\bsb{x}_1\)};
		\node[obs](x2)[below=of u1, xshift=12mm]{\(\bsb{x}_2\)};
		\node[round](u11)[above=of x1]{\(\mbf{U}_1^1\)};
		\node[round](u12)[above=of x2]{\(\mbf{U}_2^1\)};
		\node[round](f111)[above=of u1, xshift=-18mm]{\(F_{11}^1\)};
		\node[round](f112)[above=of u1, xshift=-6mm]{\(F_{12}^1\)};
		\node[round](f121)[above=of u1, xshift=6mm]{\(F_{21}^1\)};
		\node[round](f122)[above=of u1, xshift=18mm]{\(F_{22}^1\)};
		\node[round, draw=none](f1i)[above=of u1]{\(\phantom{F_{n,s}^1}\)};
		\node[round, draw=none](u2)[above=of f1i]{\(\phantom{\mbf{U}^2}\)};
		\node[round](u211)[above=of f111, xshift=1mm]{\(\mbf{U}_{11}^2\)};
		\node[round](u212)[above=of f112, xshift=1mm]{\(\mbf{U}_{12}^2\)};
		\node[round](u221)[above=of f121, xshift=1mm]{\(\mbf{U}_{21}^2\)};
		\node[round](u222)[above=of f122, xshift=1mm]{\(\mbf{U}_{22}^2\)};
		\node[round](f211)[above=of u2, xshift=-18mm]{\(F_{11}^2\)};
		\node[round](f212)[above=of u2, xshift=-6mm]{\(F_{12}^2\)};
		\node[round](f221)[above=of u2, xshift=6mm]{\(F_{21}^2\)};
		\node[round](f222)[above=of u2, xshift=18mm]{\(F_{22}^2\)};
		\node[round, draw=none](f2i)[above=of u2]{\(\phantom{F_{n,s}^2}\)};
		\node[round](f21)[above=of f2i, xshift=-12mm]{\(F_{1}^2\)};
		\node[round](f22)[above=of f2i, xshift=12mm]{\(F_{2}^2\)};
		\node[obs](y1)[above=of f21]{\(\bsb{y}_1\)};
		\node[obs](y2)[above=of f22]{\(\bsb{y}_2\)};
		\draw[-Stealth](x1)--(u11);
		\draw[-Stealth](x1.north west) to [out=120,in=270](f111.240);
		\draw[-Stealth](x1.north east) to [out=60,in=270](f112.300);
		\draw[-Stealth](x2)--(u12);
		\draw[-Stealth](x2.north west) to [out=120,in=270](f121.240);
		\draw[-Stealth](x2.north east) to [out=60,in=270](f122.300);
		\draw[-Stealth](u11)--(f111);
		\draw[-Stealth](u11)--(f112);
		\draw[-Stealth](u12)--(f121);
		\draw[-Stealth](u12)--(f122);
		\draw[-Stealth](f111.120) to [out=125,in=235](f211.240);
		\draw[-Stealth](f112.120) to [out=125,in=235](f212.240);
		\draw[-Stealth](f121.120) to [out=125,in=235](f221.240);
		\draw[-Stealth](f122.120) to [out=125,in=235](f222.240);
		\draw[-Stealth](f111.north)--(u211.south);
		\draw[-Stealth](f112.north)--(u212.south);
		\draw[-Stealth](f121.north)--(u221.south);
		\draw[-Stealth](f122.north)--(u222.south);
		\draw[-Stealth](u211.north)--(f211.south);
		\draw[-Stealth](u212.north)--(f212.south);
		\draw[-Stealth](u221.north)--(f221.south);
		\draw[-Stealth](u222.north)--(f222.south);
		\draw[-Stealth](f211)--(f21);
		\draw[-Stealth](f212)--(f21);
		\draw[-Stealth](f221)--(f22);
		\draw[-Stealth](f222)--(f22);
		\draw[-Stealth](f21)--(y1);
		\draw[-Stealth](f22)--(y2);
	\end{tikzpicture}
	\caption{
		Graphical model representations of \textit{left}: 2-layered DS-DGP, with \(S=2\) Monte Carlo samples,
		\textit{right}: AR1 for 2-layered DGP with AVI, with \(S=2\) Monte Carlo samples.
	}
	\label{fig3}
\end{figure}
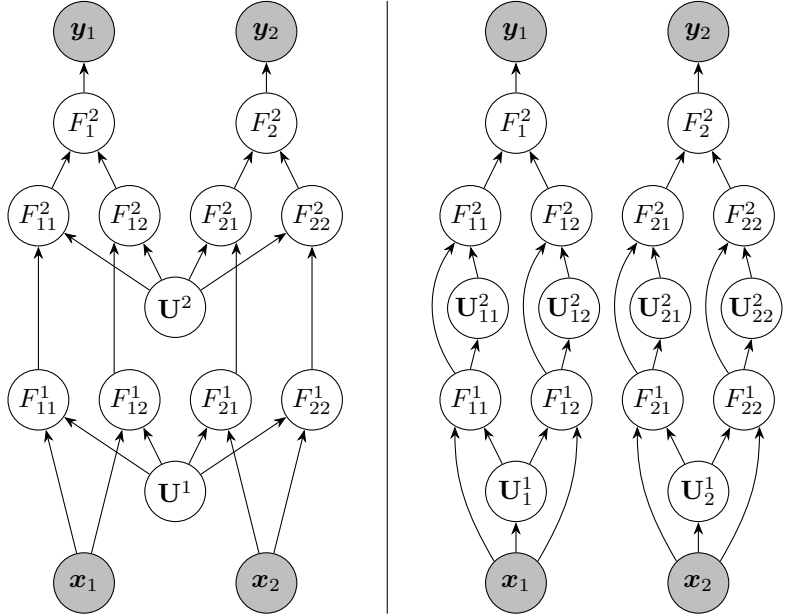

\paragraph{AR1}
For the simplest operator, i.e., the identity function \(\mcal{P}_1(F_n^{l-1})=F_n^{l-1}\),
the expected KL terms in Eq. \ref{dgp_avi_elbo} can be approximated similarly to the expected log-likelihood terms by
\begin{equation} \label{ekl_approx_1}
	\begin{split}
		\mbb{E}_{q(F_n^{l-1}|\bsb{x}_n)}\Bigl[D_\text{KL}
		\bigl(q(\mbf{U}_n^l|F_n^{l-1})\| & p(\mbf{U}_n^l|F_n^{l-1})\bigr)\Bigr] \\
		\approx\frac{1}{S}{\sum}_{s=1}^{S}D_\text{KL}\bigl(
		& q(\mbf{U}_{n,s}^l|F_{n,s}^{l-1})\|p(\mbf{U}_{n,s}^l|F_{n,s}^{l-1})\bigr)
	\end{split}
\end{equation}
except for the first layer.
\(\mbf{U}_{n,s}^l\) denotes the inducing variables conditioned on \(F_{n,s}^{l-1}\).

Intuitively, the approximation as in Eq. \ref{ekl_approx_1} uses \(S\) groups of samples
\(\{F_{n,s}^l,\mbf{U}_{n,s}^l\}_{l=1}^L\) that are lined up by random sampling between layers (Fig. \ref{fig3}, right).
Such method is straightforward and unbiased, but not unproblematic both theoretically and practically.
Firstly, the KL divergence from \(\mbf{U}_n^l\)'s variational posterior to the prior depends on
the marginal posterior \(q(F_n^{l-1}|\bsb{x}_n)\), which is a continuous Gaussian mixture.
AR1 computes separate inducing variables for each random sample \(F_{n,s}^l\),
allowing for uncertainty to be propagated through inducing variables.
Then there is a second source of uncertainty for the marginal posterior \(q(F_n^l|\bsb{x}_n)\) compared with
a DS-DGP (Fig. \ref{fig3}, left) wherein uncertainty is only propagated through layers by the inputs \(F_n^{l-1}\).
Secondly, the cost of sparse GP inference is dominated by the computations on inducing variables,
then AR1 requires \(S\) times the cost of a sparse GP for computations on all but the first layer.

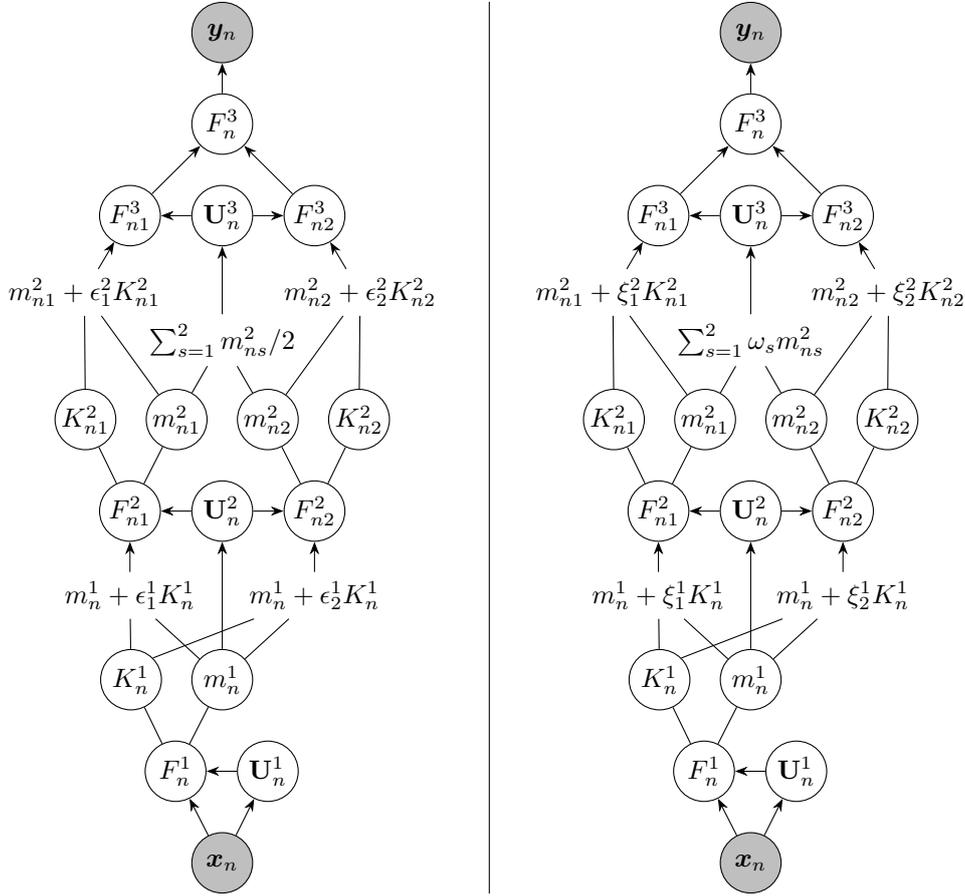
\begin{figure} [ht]
	\centering
	\begin{tikzpicture}
		[
			node distance=4mm and 4mm,
			round/.style={circle, draw=black, inner sep=0.2mm, minimum size=8mm},
			plain/.style={},
			obs/.style={circle, draw=black, fill=lightgray, inner sep=0.2mm, minimum size=8mm},
		]
		\node[round](u1){\(\mbf{U}_n^1\)};
		\node[round](F1)[left=of u1]{\(F_n^1\)};
		\node[obs](x)[below=of u1, xshift=-6mm]{\(\bsb{x}_n\)};
		\node[round, draw=none](f1i)[above=of u1, xshift=-6mm]{};
		\node[round](m1)[above=of u1, xshift=-6mm]{\(m_n^1\)};
		\node[round](K1)[above=of u1, xshift=-18mm]{\(K_n^1\)};
		\node[plain](eq1)[above=of f1i]{\(\phantom{m_n^1+\epsilon_1^1K_n^1}\)};
		\node[round](u2)[above=of eq1]{\(\mbf{U}_n^2\)};
		\node[round](F21)[left=of u2]{\(F_{n1}^2\)};
		\node[round](F22)[right=of u2]{\(F_{n2}^2\)};
		\node[plain](eq2)[below=of F21]{\(m_n^1+\epsilon_1^1K_n^1\)};
		\node[plain](eq3)[below=of F22]{\(m_n^1+\epsilon_2^1K_n^1\)};
		\node[round, draw=none](f2i)[above=of u2]{};
		\node[round](m21)[above=of u2, xshift=-6mm]{\(m_{n1}^2\)};
		\node[round](K21)[above=of u2, xshift=-18mm]{\(K_{n1}^2\)};
		\node[round](m22)[above=of u2, xshift=6mm]{\(m_{n2}^2\)};
		\node[round](K22)[above=of u2, xshift=18mm]{\(K_{n2}^2\)};
		\node[plain](eq4)[above=of f2i, yshift=-1mm]{\(\sum_{s=1}^{2}m_{ns}^2/2\)};
		\node[round](u3)[above=of eq4, yshift=5mm]{\(\mbf{U}_n^3\)};
		\node[round](F31)[left=of u3]{\(F_{n1}^3\)};
		\node[round](F32)[right=of u3]{\(F_{n2}^3\)};
		\node[plain](eq5)[below=of F31, xshift=-6mm, yshift=1mm]{\(m_{n1}^2+\epsilon_1^2K_{n1}^2\)};
		\node[plain](eq6)[below=of F32, xshift=6mm, yshift=1mm]{\(m_{n2}^2+\epsilon_2^2K_{n2}^2\)};
		\node[round](f)[above=of u3]{\(F_{n}^3\)};
		\node[obs](y)[above=of f]{\(\bsb{y}_n\)};
		\draw[-Stealth](x)--(u1);
		\draw[-Stealth](x)--(F1);
		\draw[-Stealth](u1)--(F1);
		\draw[-](F1)--(m1);
		\draw[-](F1)--(K1);
		\draw[-Stealth](m1)--(u2);
		\draw[-](m1)--(eq2);
		\draw[-](K1)--(eq2);
		\draw[-](m1)--(eq3);
		\draw[-](K1.45)--(eq3);
		\draw[-Stealth](eq2)--(F21);
		\draw[-Stealth](eq3)--(F22);
		\draw[-Stealth](u2)--(F21);
		\draw[-Stealth](u2)--(F22);
		\draw[-](F21)--(m21);
		\draw[-](F21)--(K21);
		\draw[-](F22)--(m22);
		\draw[-](F22)--(K22);
		\draw[-](m21)--(eq4);
		\draw[-](m22)--(eq4);
		\draw[-](m21)--(eq5);
		\draw[-](K21)--(eq5);
		\draw[-](m22)--(eq6);
		\draw[-](K22)--(eq6);
		\draw[-Stealth](eq4)--(u3);
		\draw[-Stealth](eq5)--(F31);
		\draw[-Stealth](eq6)--(F32);
		\draw[-Stealth](u3)--(F31);
		\draw[-Stealth](u3)--(F32);
		\draw[-Stealth](F31)--(f);
		\draw[-Stealth](F32)--(f);
		\draw[-Stealth](f)--(y);
	\end{tikzpicture}
	\hspace{4mm}
	\vrule
	\hspace{4mm}
	\begin{tikzpicture}
		[
			node distance=4mm and 4mm,
			round/.style={circle, draw=black, inner sep=0.2mm, minimum size=8mm},
			plain/.style={},
			obs/.style={circle, draw=black, fill=lightgray, inner sep=0.2mm, minimum size=8mm},
		]
		\node[round](u1){\(\mbf{U}_n^1\)};
		\node[round](F1)[left=of u1]{\(F_n^1\)};
		\node[obs](x)[below=of u1, xshift=-6mm]{\(\bsb{x}_n\)};
		\node[round, draw=none](f1i)[above=of u1, xshift=-6mm]{};
		\node[round](m1)[above=of u1, xshift=-6mm]{\(m_n^1\)};
		\node[round](K1)[above=of u1, xshift=-18mm]{\(K_n^1\)};
		\node[plain](eq1)[above=of f1i]{\(\phantom{m_n^1+\xi_1^1K_n^1}\)};
		\node[round](u2)[above=of eq1]{\(\mbf{U}_n^2\)};
		\node[round](F21)[left=of u2]{\(F_{n1}^2\)};
		\node[round](F22)[right=of u2]{\(F_{n2}^2\)};
		\node[plain](eq2)[below=of F21]{\(m_n^1+\xi_1^1K_n^1\)};
		\node[plain](eq3)[below=of F22]{\(m_n^1+\xi_2^1K_n^1\)};
		\node[round, draw=none](f2i)[above=of u2]{};
		\node[round](m21)[above=of u2, xshift=-6mm]{\(m_{n1}^2\)};
		\node[round](K21)[above=of u2, xshift=-18mm]{\(K_{n1}^2\)};
		\node[round](m22)[above=of u2, xshift=6mm]{\(m_{n2}^2\)};
		\node[round](K22)[above=of u2, xshift=18mm]{\(K_{n2}^2\)};
		\node[plain](eq4)[above=of f2i, yshift=-1mm]{\(\sum_{s=1}^{2}\omega_sm_{ns}^2\)};
		\node[round](u3)[above=of eq4, yshift=5mm]{\(\mbf{U}_n^3\)};
		\node[round](F31)[left=of u3]{\(F_{n1}^3\)};
		\node[round](F32)[right=of u3]{\(F_{n2}^3\)};
		\node[plain](eq5)[below=of F31, xshift=-6mm, yshift=1mm]{\(m_{n1}^2+\xi_1^2K_{n1}^2\)};
		\node[plain](eq6)[below=of F32, xshift=6mm, yshift=1mm]{\(m_{n2}^2+\xi_2^2K_{n2}^2\)};
		\node[round](f)[above=of u3]{\(F_{n}^3\)};
		\node[obs](y)[above=of f]{\(\bsb{y}_n\)};
		\draw[-Stealth](x)--(u1);
		\draw[-Stealth](x)--(F1);
		\draw[-Stealth](u1)--(F1);
		\draw[-](F1)--(m1);
		\draw[-](F1)--(K1);
		\draw[-Stealth](m1)--(u2);
		\draw[-](m1)--(eq2);
		\draw[-](K1)--(eq2);
		\draw[-](m1)--(eq3);
		\draw[-](K1.45)--(eq3);
		\draw[-Stealth](eq2)--(F21);
		\draw[-Stealth](eq3)--(F22);
		\draw[-Stealth](u2)--(F21);
		\draw[-Stealth](u2)--(F22);
		\draw[-](F21)--(m21);
		\draw[-](F21)--(K21);
		\draw[-](F22)--(m22);
		\draw[-](F22)--(K22);
		\draw[-](m21)--(eq4);
		\draw[-](m22)--(eq4);
		\draw[-](m21)--(eq5);
		\draw[-](K21)--(eq5);
		\draw[-](m22)--(eq6);
		\draw[-](K22)--(eq6);
		\draw[-Stealth](eq4)--(u3);
		\draw[-Stealth](eq5)--(F31);
		\draw[-Stealth](eq6)--(F32);
		\draw[-Stealth](u3)--(F31);
		\draw[-Stealth](u3)--(F32);
		\draw[-Stealth](F31)--(f);
		\draw[-Stealth](F32)--(f);
		\draw[-Stealth](f)--(y);
	\end{tikzpicture}
	\caption{
		Graphical model representations of \textit{left}: AR2 for 3-layered DGP with AVI,
		with \(S=2\) Monte Carlo samples,
		\textit{right}: AR2P for 3-layered DGP with AVI, with \(S=2\) quadrature points.
	}
	\label{fig4}
\end{figure}

\paragraph{AR2}
A more efficient amortization rule (Fig. \ref{fig4}, left) is to take the means of the latent function values as inputs,
i.e., \(\mcal{P}_2(F_n^{l-1})=\mbb{E}[F_n^{l-1}]\).
It is mentioned above that \(q(F_n^{l-1}|\bsb{x}_n)\) is a continuous Gaussian mixture for \(l\geq 3\),
of which the mean can be approximated by
\begin{equation}
	\begin{split}
		\mbb{E}_{q(F_n^{l-1}|\bsb{x}_n)}[F_n^{l-1}]
		& =\int F_n^{l-1}q(F_n^{l-1}|F_n^{l-2})\,dF_n^{l-1}{\prod}_{i=1}^{l-2}q(F_n^i|F_n^{i-1})\,dF_n^i \\
		& \approx\frac{1}{S}{\sum}_{s=1}^{S}\mbb{E}_{q(F_{n,s}^{l-1}|F_{n,s}^{l-2})}[F_{n,s}^{l-1}]
		=\frac{1}{S}{\sum}_{s=1}^{S}m_{n,s}^{l-1}=\hat{m}_{n,(S)}^{l-1},
	\end{split}
\end{equation}
and the second layer has \(\hat{m}_{n,(S)}^1=m_n^1\).
In this setting the random samples \(\{F_{n,s}^l\}_{s=1}^S\) share the same inducing variables \(\mbf{U}_n^l\).
The expected KL terms in Eq. \ref{dgp_avi_elbo} are approximated by
\begin{equation} \label{ekl_approx_2}
	\mbb{E}_{q(F_n^{l-1}|\bsb{x}_n)}\Bigl[
		D_\text{KL}\bigl(q(\mbf{U}_n^l|F_n^{l-1})\|p(\mbf{U}_n^l|F_n^{l-1})\bigr)
	\Bigr]\approx D_\text{KL}\bigl(
		q(\mbf{U}_n^l|\hat{m}_{n,(S)}^{l-1})\|p(\mbf{U}_n^l|\hat{m}_{n,(S)}^{l-1})
	\bigr)
\end{equation}
except for the first layer.

AR2 reduces the uncertainty passed to inducing variables by avoiding direct sampling from the latent function values.
Moreover, the computational cost of each layer is recovered to that of a sparse GP
(see Appendix \ref{App_compute_cost} for details).
Importantly, one can verify that both AR1 and AR2 are consistent with the first layer:
Random samples (AR1) from a Gaussian density with zero variance are always equal to the mean (AR2),
which is exactly the input \(\bsb{x}_n\).

\subsection{A method of determinism} \label{Sec_ARP}

As mentioned above, the non-injective nature of GP mappings makes DGPs susceptible to pathologies.
The posterior approximation in the doubly stochastic formulation aggravates the problem,
since random sampling brings uncertainty between layers.
This is potentially problematic for our method as the uncertainty of latent function values
might degrade the performance of amortization.
Here we leverage the deep sigma point process (DSPP, \citet{pmlr-v124-jankowiak20a}) to construct
a fully parameterized formulation of model for AR2, to which we refer as AR2P (i.e., AR2 parametric).

\paragraph{AR2P}
In the doubly stochastic formulation (Fig. \ref{fig4}, left), the inputs of the \(l\)-th layer are sampled from
\(F_{n,s}^{l-1}\sim\mcal{N}(m_{n,s}^{l-1},K_{n,s}^{l-1})\) as \(m_{n,s}^{l-1}+\epsilon_s^{l-1}K_{n,s}^{l-1}\),
where \(\epsilon_s^{l-1}\sim\mcal{N}(0,1)\).
AR2P (Fig. \ref{fig4}, right) instead parameterizes the variational posteriors with learnable quadratures
\citep{pmlr-v124-jankowiak20a}. To be specific, learnable quadrature points \(\{\xi_s^l\}_{l=1,s=1}^{L-1,S}\)
are introduced to replace random numbers \(\{\epsilon_s^l\}_{l=1,s=1}^{L-1,S}\), with the resulting \(S\)-component
Gaussian mixtures controlled by learnable quadrature weights \(\{\omega_s\}_{s=1}^S\) that sum to unity.
The marginal posterior at the \(l\)-th layer (\(l\geq 2\)) is given by finite Gaussian mixture
\begin{equation}
	q(F_n^l|\bsb{x}_n)={\sum}_{s=1}^{S}\omega_sq(F_{n,s}^l|F_{n,s}^{l-1}),
\end{equation}
with mean \(\mbb{E}_{q(F_n^l|\bsb{x}_n)}[F_n^l]=\sum_{s=1}^{S}\omega_sm_{n,s}^l\) serves as inputs to
the inference function of the subsequent layer.
The expected log-likelihood terms in Eq. \ref{dgp_avi_elbo} are parameterized by
\begin{equation} \label{ell_approx_2}
	\mbb{E}_{q(F_n^L|\bsb{x}_n)}\bigl[\ln p(\bsb{y}_n|F_n^L)\bigr]={\sum}_{s=1}^{S}
	\omega_s\mbb{E}_{q(F_{n,s}^L|F_{n,s}^{L-1})}\bigl[\ln p(\bsb{y}_n|F_{n,s}^L)\bigr].
\end{equation}

The difference between AR2P and AR2 lies in the treatment of latent function values.
Remarkably, the marginal posteriors depend on \(S\) deterministic quadrature-dependent feature sets
under such parameterization, i.e., the uncertainty represented by variances \(\{K_{n,s}^l\}_{l=1,s=1}^{L,S}\)
is propagated through layers in a deterministic regime instead of random sampling.

\subsection{Predictions}

To predict on a test point \(\bsb{x}_*\) we sample from the marginal posterior at the final layer,
\begin{equation}
	q(F_*^L|\bsb{x}_*)=\int q(F_*^L|F_*^{L-1}){\prod}_{l=1}^{L-1}q(F_*^l|F_*^{l-1})\,dF_*^l,
\end{equation}
whose density is approximated by a finite Gaussian mixture
\begin{equation}
	q(F_*^L|\bsb{x}_*)\approx\frac{1}{S}{\sum}_{s=1}^Sq(F_{*s}^L|F_{*s}^{L-1})
\end{equation}
with mean \(\hat{m}_{*(S)}^L\) for AR1 and AR2, where we draw \(S\) samples \(\{F_{*s}^{L-1}\}_{s=1}^S\)
from the marginal posterior at the penultimate layer.

For AR2P, the marginal posterior is parameterized by weighted Gaussian mixture
\begin{equation}
	q(F_*^L|\bsb{x}_*)={\sum}_{s=1}^{S}\omega_sq(F_{*s}^L|F_{*s}^{L-1})
\end{equation}
with mean \(\sum_{s=1}^{S}\omega_sm_{*s}^L\).

\subsection{Further model details} \label{Sec_detail}

\paragraph{Mean function}
As zero mean functions are considered inappropriate for the intermediate layers of DGPs, \citep{pmlr-v33-duvenaud14}
partially solved the problem by concatenating the inputs with the outputs at each GP layer.
\citet{NIPS2017_82089746} instead suggested to include linear mean functions
\(m^l(\mbf{F}^{l-1})=\mbf{F}^{l-1}\mbf{W}^l\) derived by PCA for \(l<L\).
As the complexity of PCA is cubic in the input dimension \(D^l\),
we consider an efficient alternative with learnable weights \(\mbf{W}^l\) for the mean function of each layer,
though unnecessary in our method.

\paragraph{Variational approximation}
The approximate covariance matrices \(\bsb{\Sigma}\) are usually specified by Cholesky factors \(\mbf{L}\)
with quadratic \(M(M+1)/2\) parameters. We deem such parameterization costly in deep models and suggest
to restrict \(\bsb{\Sigma}_n^l\) to be diagonal instead, i.e., the mean-field variational approximation.

\paragraph{Training objective}
\citet{pmlr-v119-jankowiak20a, pmlr-v124-jankowiak20a} used a modified training objective that directly targets
the predictive distribution for the regression tasks with Gaussian likelihood (see Appendix \ref{App_pp}).
A similar modification can be applied to AR2P of our method, with the expected log-likelihood terms as in
Eq. \ref{ell_approx_2} replaced by
\begin{equation}
	\ln p(\bsb{y}_n|\bsb{x}_n)=\ln{\sum}_{s=1}^{S}\omega_s
	\mcal{N}\bigl(\bsb{y}_n|m_{n,s}^L,K_{n,s}^L+\sigma_{obs}^2\mbf{I}\bigr).
\end{equation}
We denote the learnable quadrature-parameterized model with AR2 by AR2P in Sec. \ref{Sec_ARP},
where P signifies parametric.
When AR2P is trained with the modified objective function,
we denote such method by AR2PP (i.e., AR2 parametric predictive).

\paragraph{Model initialization}
In sparse GPs with SVI the inducing points are often initialized by K-means or PCA,
with approximate mean initialized to zero, approximate covariance initialized to identity.
Since the inducing points are input dependent in our method,
we initialize the \(M\) affine functions with identity weight matrices \(\mbf{W}_m\)'s
and random biases \(\mbf{b}_m\)'s s.t. the initial inducing points are noisy corruptions of the input.
The inference functions of approximate means and covariances are initialized s.t.
the initial outputs over the \(M\) dimensions are all equal.

\section{Related work}

\citet{NIPS2005_4491777b} firstly introduced inducing point techniques for approximate GP inference and
computed them with maximum likelihood methods, resulting in an inaccurate posterior compared with exact GPs.
\citet{pmlr-v5-titsias09a} considered a variational approach wherein the inducing points
are treated as parameters and determined by optimizing a lower bound.
\citet{10.5555/3023638.3023667} introduced SVGP by combining SVI \citep{JMLR:v14:hoffman13a} with sparse GPs.
SVGP improved the lower bound in \citet{pmlr-v5-titsias09a} by variationally integrating out inducing variables,
which enables the lower bound to factorize across datapoints.
\citet{pmlr-v162-jafrasteh22a} innovatively applied AVI \citep{margossian2024amortized} to sparse GPs,
i.e., to map the inputs to variational parameters with an inference function.
The variational posterior of \citet{pmlr-v162-jafrasteh22a}  is more expressive than that of SVGP
as it uses separate input dependent inducing variables for conditioning.
This method also performs well with a fairly small number \(M\) of inducing points,
thereby further improving the computational cost.

Modern multilayer stacked DGPs were formally introduced by \citet{pmlr-v31-damianou13a},
with mean-field variational posteriors that assume independence and Gaussianity between layers.
\citet{NIPS2017_82089746} proposed a more flexible architecture that retains the exact model posterior
while maintaining the correlations within and across adjacent layers.
They introduced a doubly-stochastic variational approach that combines Monte Carlo methods
to construct a tractable lower bound.
The predictive distribution in \citet{NIPS2017_82089746} is a continuous Gaussian mixture
and in practice approximated with a finite Gaussian mixture obtained by random sampling.
\citet{pmlr-v124-jankowiak20a} replaced the random samples between layers with learnable quadratures
(or sigma point approximations) to construct a completely parameterized model
to which they referred as deep sigma point process (DSPP).
DSPPs are typically trained with a modified lower bound that directly targets the predictive distribution
\citep{pmlr-v119-jankowiak20a}, which can be viewed as a maximum likelihood estimation (MLE) of the parametric model.

Our method directly builds on \citet{NIPS2017_82089746} and \citet{pmlr-v162-jafrasteh22a} and
incorporates the parametric class of GP models \citet{pmlr-v119-jankowiak20a, pmlr-v124-jankowiak20a}.
Specifically, we improve the amortization function in \citet{pmlr-v162-jafrasteh22a}
and extend the idea of AVI \citet{margossian2024amortized} to DGPs
by successively amortizing variational parameters in the GP layers.
We reformulate the compositional variational posterior with corresponding variational lower bound
and introduce strategies for tractable approximate inference.

\section{Empirical evaluation}

In this section we explore the empirical performance of DGPs with AVI as introduced in Sec. \ref{Sec_DGP_AVI}.
For simplicity, we refer to the proposed method as \emph{amortized variational deep Gaussian process} (AVDGP).
We also consider two shallow models, i.e., SOLVE-GP \citep{pmlr-v108-shi20b}, IDSGP \citep{pmlr-v162-jafrasteh22a},
and two deep models, i.e., DS-DGP \citep{NIPS2017_82089746}, DSPP \citep{pmlr-v124-jankowiak20a} as baselines.
For details about mean functions, kernels, numbers of inducing variables, etc., refer to Appendix \ref{App_train_conf}.

\subsection{Amortization rule ablation study} \label{Sec_ar}

We commence with an investigation on the three amortization rules proposed in Sec. \ref{Sec_AR} and Sec. \ref{Sec_ARP}.
In particular 3-layered models are used for comparison as AR2 and AR2P
differs in the inputs of inference functions from the third layer on.
For AR1 we set \(S=1\) at training time, and \(S=32\) at validation/test time (see Appendix \ref{App_compute_cost}).
For AR2 and AR2P we set \(S=32\).
The models are trained on 8 smallest regression datasets described in Sec. \ref{Sec_regr}
with sample size in the range \(5\times 10^3\lessapprox N\lessapprox 5\times 10^4\).
The results are summarized in Tab. \ref{tab1}, with more details in Appendix \ref{App_ar}.
Unsurprisingly, the three amortization rules show largely comparable performance,
with some preference for AR2P that behaves slightly better.
Nevertheless, AR2 and AR2P are considered to be more efficient as they cost less in making predictions.
We use AR2P in the remainder of the experiments in preference to deterministic outputs and thus reproducible results.

\begin{table} [ht]
	\centering
	\caption{
		Average rankings and standard deviations (over 5 optimization runs) of proposed amortization rules
		on 8 univariate regression datasets.
		NLL: negative log-likelihood; RMSE: root mean squared error;
		CRPS: continuous ranked probability score.
	}
    \begin{tblr}{
		cells={c,m},
        hline{1,Z}={0.75pt},
		hline{2},
    }
	& AR1 & AR2 & AR2P \\
	NLL & 2.23\,(0.49) & 2.03\,(0.51) & 1.75\,(0.43) \\
	RMSE & 2.18\,(0.65) & 1.93\,(0.67) & 1.90\,(0.57) \\
	CRPS & 2.08\,(0.63) & 2.25\,(0.68) & 1.68\,(0.50)
    \end{tblr}
    \label{tab1}
\end{table}

\subsection{Toy problem}

\begin{wrapfigure}{r}{6cm}
	\includegraphics[width=5cm, center]{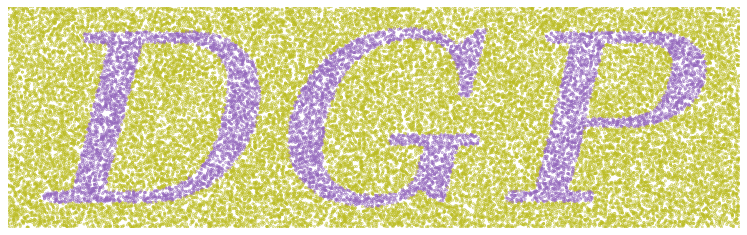}
	\captionsetup{width=5.5cm}
	\caption{Visualization of training data (\(N=40000\)). \(y=1\) for purple points; \(y=0\) for olive points.}
	\label{dgp_image}
\end{wrapfigure}

We illustrate our method with a binary classfication problem on 2D boundaries with complex non-Gaussian density.
The data is randomly generated from \(X_1,X_2\sim U(0,1)\), with label \(y\) indicates whether
\((X_1,X_2)\) is located inside the letters \enquote{DGP} (Fig. \ref{dgp_image}).

We investigate how the representational power of AVDGP can be impacted by the number \(|M|\) of inducing points
and the number \(L\) of GP layers, where we define \(|M|=\sum_{l=1}^{L}M^l\).
We fit AVDGPs with \(|M|=[16,32,64]\) and \(L=[1,2,3,4]\).
Notably, when \(L=1\) the model reduces to a shallow GP with inference functions as in Eq. \ref{affine_func},
to which we refer as SGP-AVI for clarity. We set the proportion of inducing variables in each layer to
1:1 for 2-layered models, 2:1:1 for 3-layered models, and 4:2:1:1 for 4-layered models.
The test error rates are shown in Tab. \ref{tab2}. Increasing the number of inducing points does improve the
performance, but the information communicated by inducing points reaches a saturation when \(|M|\) is large enough.
We also observe that the model performs better with increased depth,
which corresponds with previous works on DGPs, i.e., the expressivity of DGPs significantly depends on
the number of GP layers, as deeper models have the potential to represent more complicated functions.
Therefore, we suggest to increase the depth of AVDGPs rather than solely increase the number of inducing points
when modeling very complex data.

\begin{table} [ht]
	\centering
	\caption{
		Average test error rates and standard deviations (over 5 optimization runs) on the toy classfication problem.
	}
    \begin{tblr}{
		cells={c,m},
        hline{1,Z}={0.75pt},
		hline{2},
		vline{2},
    }
	\(|M|\) & SGP-AVI & AVDGP-2L & AVDGP-3L & AVDGP-4L \\
	16 & 10.41\%\,(1.54\%) & 8.02\%\,(0.70\%) & 4.51\%\,(1.13\%) & 4.94\%\,(1.69\%) \\
	32 & 7.63\%\,(1.43\%) & 4.40\%\,(1.25\%) & 4.14\%\,(0.68\%) & 2.59\%\,(0.27\%) \\
	64 & 6.81\%\,(1.74\%) & 3.10\%\,(0.48\%) & 2.59\%\,(0.44\%) & 1.86\%\,(0.28\%) \\
	128 & 6.04\%\,(1.02\%) & 2.92\%\,(0.69\%) & 3.03\%\,(0.59\%) & 2.43\%\,(1.08\%)
    \end{tblr}
    \label{tab2}
\end{table}

\subsection{Benchmark datasets} \label{Sec_regr}

We consider 12 univariate regression datasets extracted from the UCI repository \citep{UCIrepository},
with two additional benchmarks, i.e., Kin8nm and Kin32nm, from the Kin family of datasets\footnote{
	\url{https://www.cs.toronto.edu/~delve/data/kin/desc.html}
}.
Some existing methods are used for comparison with our method, i.e., SOLVE,
a sparse GP based on an orthogonal decomposition that allows introduction of an additional set of inducing points,
DS-DGP as described in Sec. \ref{Sec_DGP}, DSPP as described in Sec. \ref{Sec_ARP},
and IDSGP as described in Sec. \ref{Sec_SGP_AVI}.
We focus on three calibration metrics, i.e., negative log-likelihood (NLL), root mean squared error (RMSE) and
continuous ranked probability score (CRPS, \citet{doi:10.1198/016214506000001437}, see Appendix \ref{App_crps}).
The results are summarized in Tab. \ref{tab3} and Figs. \ref{regr_nll}-\ref{regr_crps},
with more details in Appendix \ref{App_regr}.

\begin{table} [ht]
	\centering
	\caption{
		Average rankings and standard deviations (over 5 optimization runs) of the methods
		in Sec. \ref{Sec_regr} on 14 univariate regression datasets.
	}
    \begin{tblr}{
		cells={c,m},
        hline{1,Z}={0.75pt},
		hline{2},
		vline{X}={dotted},
	}
	& SOLVE & DS-DGP & DSPP & IDSGP & AR2P & AR2PP \\
	NLL  & 5.63\,(0.16) & 4.80\,(0.32) & 2.19\,(0.18) & 4.09\,(0.29) & 3.03\,(0.28) & \tbf{1.27}\,(0.18) \\
	RMSE & 5.19\,(0.35) & 4.20\,(0.54) & 3.66\,(0.48) & 3.00\,(0.45) & \tbf{2.13}\,(0.54) & 2.83\,(0.65) \\
	CRPS & 5.74\,(0.08) & 4.73\,(0.27) & 2.79\,(0.21) & 4.04\,(0.42) & 2.41\,(0.31) & \tbf{1.29}\,(0.23)
    \end{tblr}
    \label{tab3}
\end{table}

In aggregate AVDGP-AR2P outperforms all competitors in terms of RMSE and CRPS.
It also performs better than the other three methods expect DSPP in terms of NLL.
While DSPP achieves the best NLL on most datasets,
we ascribe this to its modified training objective \citep{pmlr-v119-jankowiak20a, pmlr-v124-jankowiak20a}
that tends to underestimate the observation noise \(\sigma_{obs}^2\)
and sacrifice the accuracy of point estimate for better predictive performance.
We notice that, shallow model notwithstanding,
IDSGP outperforms DS-DGP in terms of all metrics and DSPP in terms of RMSE.
We attribute its high capability to the modified structure with local inducing variables to each input.
Since AVDGP is in effect a deep hierarchy composed of IDSGP layers, it yields even better performance.

We also consider AR2PP for our method as in Sec. \ref{Sec_detail}, with the results also included
in Tab. \ref{tab3} and Figs. \ref{regr_nll}-\ref{regr_crps}.
AVDGP-AR2PP outperforms AVDGP-AR2P in terms of NLL and CRPS,
but we observe that the modified training objective can degrade the RMSE performance as argued above.
Nevertheless, AVDGP-AR2PP is superior to DSPP, its SVI counterpart, in terms of all metrics.

\begin{figure} [ht]
	\includegraphics[width=17cm, center]{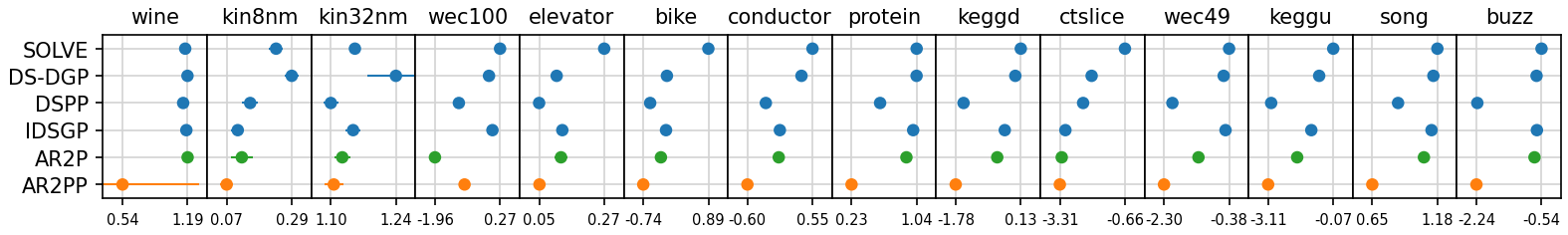}
	\caption{
		Test negative log-likelihood (NLL) on univariate regression datasets
		(further to the left is better). Results are averaged over 5 random train/test/validation splits.
		Here and throughout uncertainty bars depict standard deviations.
	}
	\label{regr_nll}
\end{figure}

\begin{figure} [ht]
	\includegraphics[width=17cm, center]{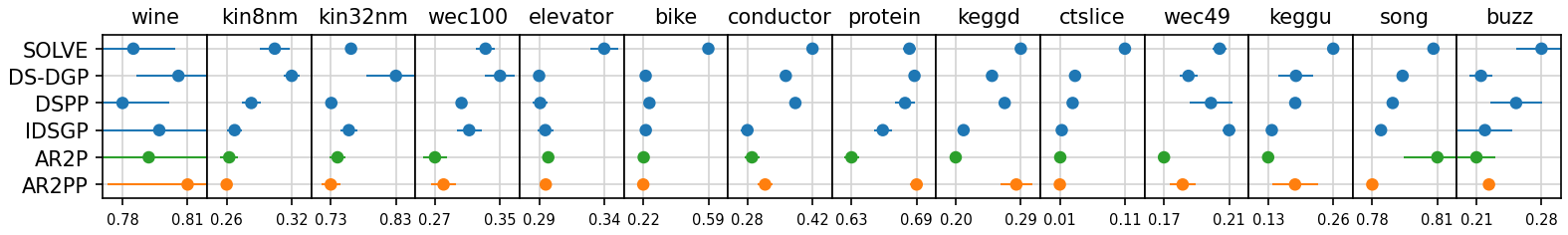}
	\caption{
		Test root mean squared error (RMSE) on univariate regression datasets
		(further to the left is better). Results are averaged over 5 random train/test/validation splits.
	}
	\label{regr_rmse}
\end{figure}

\begin{figure} [ht]
	\includegraphics[width=17cm, center]{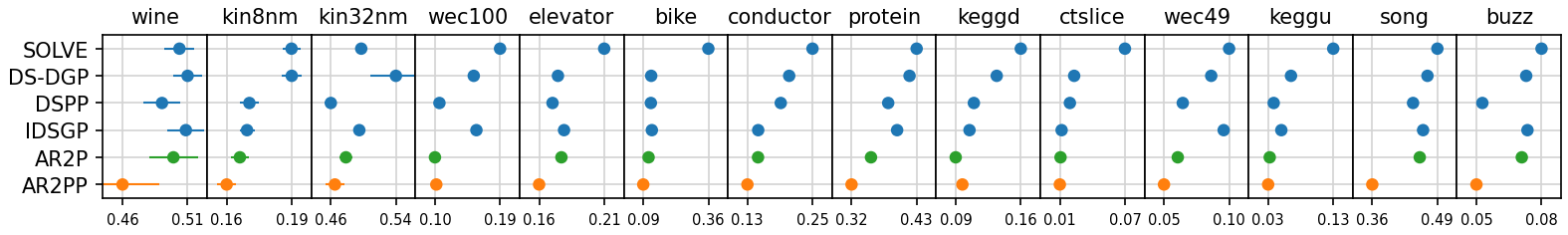}
	\caption{
		Test continuous ranked probability score (CRPS)
		on univariate regression datasets (further to the left is better).
		Results are averaged over 5 random train/test/validation splits.
	}
	\label{regr_crps}
\end{figure}

DGPs are notoriously difficult to train due to the multilayer structure and complex intra-layer computations.
This computational burden is considerably alleviated by AVDGP,
which drastically reduces the number of inducing points involved.
We measure the average training time and test time on an NVIDIA GeForce RTX 3090 GPU for the three deep models,
with results presented in Appendix \ref{App_regr}.
As detailed in Appendix \ref{App_compute_cost}, most computational cost of AVDGP lies in
the amortization procedure, which is roughly quadratic in the dimensions \(\{D^l\}_{l=0}^{L-1}\).
So AVDGP generally runs faster on low-dimensional data while costing more on high-dimensional data.
The cost of sparse GP models with SVI is dominated by \(\mcal{O}(M^3)\) that is free of batch size \(B\).
In contrast, the complexity grows linearly with \(B\) in AVDGP as all computations are input dependent,
then increasing the batch size cannot shorten the training time of an epoch.

\section{Conclusion}

The expressivity of shallow sparse GPs are limited,
as not all prior knowledge can express solely via means and kernels,
and the marginal posterior with strong Gaussianity assumption is insufficient for complex functions.
DGPs overcome the limitations by parameterizing multiple GP mappings to learn
low-dimensional embeddings of inputs and construct non-Gaussian marginal posteriors.
However, DGPs suffer from potential prior collapse, i.e., the compositional prior loses representational ability
as the number of layers increases, as well as poor scalability incurred by the hierarchical architecture.
In this work we address the deficiencies in previous DGP models by successively amortizing variational parameters
for the sparse GP layers to construct an input dependent prior and corresponding marginal posterior.
As the amortized inducing points are more informative than those in conventional sparse GPs,
our method has a more expressive prior conditioned on fewer inducing points.
Importantly, the modified prior does not degenerate with more layers.
We then exploit the compositional structure of prior to build a novel amortized approximate posterior
of high capability and formulate a tractable lower bound for optimization.
Empirical investigations demonstrate that our method behaves well on complex synthetic and real-world datasets
and outperforms a number of strong baselines at a smaller computational cost.

\clearpage
\pagestyle{plain}
\newgeometry{margin=2cm, headsep=0.5cm, footskip=1cm}
\bibliographystyle{hunsrtnat}
\bibliography{iddgp}

\clearpage
\pagestyle{fancy}
\fancyhead{}
\fancyhead[C]{\tbf{Amortized Variational Inference for Deep Gaussian Processes}}
\newgeometry{margin=3cm, headsep=0.5cm, footskip=1cm}
\appendix

\section{Derivation of the ELBO} \label{derive_dselbo}

Consider a meta-point \(\bsb{x}_*\) that determines the variational parameters of inducing variables
\(\{\mbf{U}_*^l\}_{l=1}^L\), then the extended approximate posterior takes the form
\begin{equation*}
	q\bigl(\{\mbf{F}^l,\mbf{U}_*^l\}_{l=1}^L,\bsb{x}_*\bigr)
	=p(\bsb{x}_*)\prod_{l=1}^{L}p(\mbf{F}^l|\mbf{U}_*^l,\mbf{F}^{l-1})q(\mbf{U}_*^l|F_*^{l-1}),
\end{equation*}
where \(p(\bsb{x}_*)\) is the distribution over data.
We lower bound the log marginal likelihood as
\begin{equation*}
	\begin{split}
		\ln p(\mbf{Y}|\mbf{X})\geq\mcal{L}
		& =\mbb{E}_{q(\{\mbf{F}^l,\mbf{U}_*^l\}_{l=1}^L,\bsb{x}_*)}\left[
			\ln\frac{p(\mbf{Y}|\mbf{F}^L){\color{gray}p(\bsb{x}_*)}\prod_{l=1}^{L}
			{\color{gray}p(\mbf{F}^l|\mbf{U}_*^l,\mbf{F}^{l-1})}p(\mbf{U}_*^l|F_*^{l-1})}
			{{\color{gray}p(\bsb{x}_*)}\prod_{l=1}^{L}
			{\color{gray}p(\mbf{F}^l|\mbf{U}_*^l,\mbf{F}^{l-1})}q(\mbf{U}_*^l|F_*^{l-1})}
		\right] \\
		& =\mbb{E}_{p(\bsb{x}_*)}\biggl[
			{\sum}_{n=1}^{N}\mbb{E}_{q(F_n^L|\bsb{x}_n,\bsb{x}_*)}\bigl[\ln p(\bsb{y}_n|F_n^L)\bigr] \\
			& \hspace{5em}-{\sum}_{l=1}^{L}\mbb{E}_{q(F_*^{l-1}|\bsb{x}_*)}
			\Bigl[D_\text{KL}\bigl(q(\mbf{U}_*^l|F_*^{l-1})\|p(\mbf{U}_*^l|F_*^{l-1})\bigr)\Bigr]
		\biggr].
	\end{split}
\end{equation*}
The outermost expectation w.r.t. the implicit data distribution \(p(\cdot)\) is generally analytically intractable,
but direct to random sampling.
Since samples \(\mbf{X}\) follow the data distribution, i.e., \(\bsb{x}_n\sim p(\bsb{x}_n)\),
the lower bound can be approximated by
\begin{equation*}
	\begin{split}
		\mcal{L}\approx{\sum}_{n=1}^N\biggl\{
			& \mbb{E}_{q(F_n^L|\bsb{x}_n)}\bigl[\ln p(\bsb{y}_n|F_n^L)\bigr] \\
			& -\frac{1}{N}{\sum}_{l=1}^L\mbb{E}_{q(F_n^{l-1}|\bsb{x}_n)}\Bigl[
			D_\text{KL}\bigl(q(\mbf{U}_n^l|F_n^{l-1})\|p(\mbf{U}_n^l|F_n^{l-1})\bigr)\Bigr]\biggr\}.
	\end{split}
\end{equation*}
Furthermore, when we train with stochastic gradient optimizers,
an estimate of the lower bound on random mini-batch \(\{\bsb{x}_b,\bsb{y}_b\}_{b=1}^B\) is
\begin{equation*}
	\begin{split}
		\hat{\mcal{L}}={\sum}_{b=1}^B\biggl\{
			& \frac{N}{B}\mbb{E}_{q(F_b^L|\bsb{x}_b)}\bigl[\ln p(\bsb{y}_b|F_b^L)\bigr] \\
			& -\frac{1}{B}{\sum}_{l=1}^L\mbb{E}_{q(F_b^{l-1}|\bsb{x}_b)}\Bigl[
			D_\text{KL}\bigl(q(\mbf{U}_b^l|F_b^{l-1})\|p(\mbf{U}_b^l|F_b^{l-1})\bigr)\Bigr]\biggr\}.
	\end{split}
\end{equation*}

\section{Time and space complexity} \label{App_compute_cost}

Single-output sparse GPs typically have cubic computational cost \(\mcal{O}(M^3)\) and quadratic memory
requirement \(\mcal{O}(M^2)\), then the \(l\)-th GP layer have \(D^l\) times the time and space complexity.
The complexity of amortizing the variational parameters with inference function
Eq. \ref{infer_func} for the \(l\)-th layer is roughly \(\mcal{O}(M(D^{l-1})^2+|\phi_l|+|\psi_l|)\),
where \(|\phi_l|\) and \(|\psi_l|\) denote the number of parameters in neural networks.
The computations for each layer are done in order, so AR2 and AR2P has time complexity
\begin{equation*}
	{\sum}_{l=1}^{L}\mcal{O}\bigl((M^l)^3D^l+M^l(D^{l-1})^2+|\phi_l|+|\psi_l|\bigr)
\end{equation*}
and space complexity
\begin{equation}
	{\sum}_{l=1}^{L}\mcal{O}\bigl((M^l)^2D^l+M^l(D^{l-1})^2+|\phi_l|+|\psi_l|\bigr).
\end{equation}
AR1 requires \(S\) times the time complexity for all but the first layer,
and \(S\) times the space complexity for the computations on GPs.
Remark that for AR1 \(S\) can be as small as \(1\) when we train with stochastic gradient optimizers.
However, at test time a large \(S\) is preferable for accurate predictions, which is very expensive for AR1.

\section{Further experiments \& Experimental details}

\subsection{Training details} \label{App_train_conf}

Throughout the models we use isotropic Mat\'{e}rn-\(5/2\) kernels.
Shallow GPs and our method have zero mean functions,
while DS-DGP and DSPP have linear mean functions with learned weights for intermediate layers.
Shallow GPs use full-rank variational approximations, while deep models use mean-field variational approximations.
For the deep models we consider 3-layered structure with \(D^1=16\), \(D^2=4\) in regression tasks,
and \(D^l=2\) for all intermediate layers in the toy classfication problem unless otherwise stated.
We use the simplest multilayer perceptrons (MLPs) with 2 hidden layers as amortization neural networks
in IDSGP and AVDGP. The widths of hidden layers are the smaller of input/output dimensions.
The nonlinearities between layers are set to be LeakyReLU with negative slope \(\alpha=0.2\).

In the regression settings, we use the Adam optimizer with initial learning rate \(\ell=0.005\).
For the two largest datasets, i.e., Song and Buzz, we set mini-batch size \(B=500\) and train for 50 epochs.
For the other datasets, we set \(B=100\) and train for 100 epochs.
We do 5 train/test/validation splits on all datasets, always in the proportion 8:1:1.
All datasets are standardized in both input and output space. SOLVE-GP uses \(M_1=M_2=512\) inducing points.
DS-DGP uses \(M^1=256\), \(M^2=M^3=128\) inducing points for the 3 layers.
In IDSGP we set \(M=16\), and in AVDGP \(M^1=8\), \(M^2=M^3=4\).
We use \(S=32\) Monte Carlo samples for approximation in DS-DGP,
while for AVDGP-AR2P we use \(S=32\) quadrature points.
In the toy classfication problem, we use the same optimizer,
mini-batch size and number of epochs as in the regression tasks.
We hold out a test set and do 5 random train/validation splits, with train/test/validation proportion 3:2:1.

\subsection{Inference function ablation study} \label{App_infer_func}

\begin{figure} [ht]
	\includegraphics[width=12cm, center]{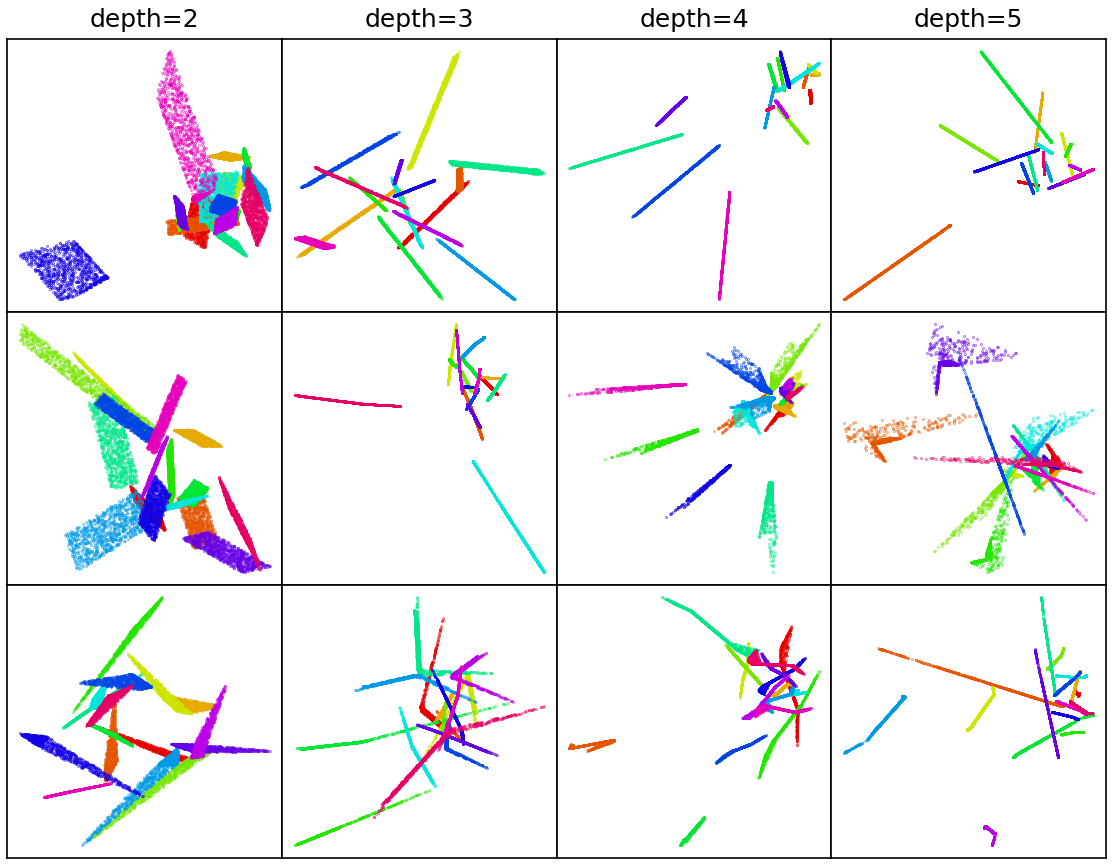}
	\caption{
		Visualization of the locations of inducing points after optimization.
		\(M=8\) groups of inducing points are represented by differet colors.
		We choose 3 models with smallest test error rates for visualization from 5 optimization runs.
	}
	\label{toy_induc}
\end{figure}

\begin{figure} [ht]
	\includegraphics[width=12cm, center]{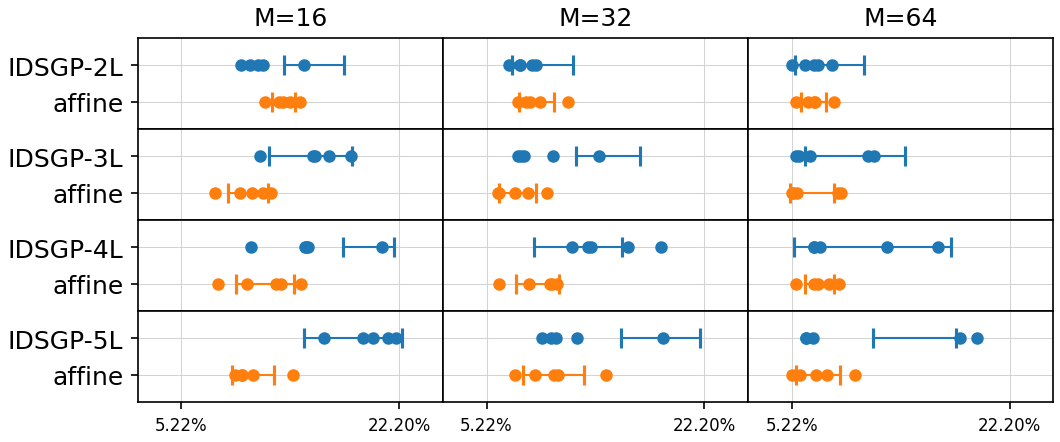}
	\caption{
		Test error rates of IDSGPs on toy dataset. Each model is optimized for 5 times.
		Points denote optimization runs, and error bars depict means \(\pm\) stds.
		The results become unstable as the depth of amortization neural networks increases,
		i.e., some models fail to converge.
		The problem is eased by using affine transformations as inference functions.
	}
	\label{toy_affine}
\end{figure}

We argued that IDSGP with deep amortization neural networks can suffer from pathologies in Sec. \ref{Sec_affine}.
Again, we illustrate this problem by experiments on the toy dataset.
As the depth of amortization neural networks increases, the inducing points tend to be located in
a subspace of the input space \(\mcal{X}\) (Fig. \ref{toy_induc}). The expressivity of resulting GP prior
can be severely degraded, obstructing the the model from convergence (Fig. \ref{toy_affine}).

The inference function as in Eq. \ref{affine_func} instead maps inputs to inducing points with learned
affine transformations. Note that an affine transformation maps a space to itself, then the inducing points
can hardly degenerate compared with the neural networks used by \citet{pmlr-v162-jafrasteh22a}.
An ablation study on the toy problem confirms above theoretical reasoning, as shown in Fig. \ref{toy_affine}.
The proposed inference function achieves similar or even better performance to the original
amortization neural networks with much smaller standard deviations across optimization runs.

\subsection{Amortization rule ablation study} \label{App_ar}

We include the detailed results of Sec. \ref{Sec_ar} in Tab. \ref{tab5}.

\subsection{Benchmark datasets} \label{App_regr}

We include the detailed results of Sec. \ref{Sec_regr} in Tab. \ref{tab5}, with train/test time in Tab. \ref{tab4}.

\begin{table} [ht]
	\footnotesize
	\centering
	\caption{
		Average training and test time per epoch with standard deviations (over 5 optimization runs)
		of the deep models in Sec. \ref{Sec_regr} on regression datasets.
	}
    \begin{tblr}{
		cells={c,m},
		columns={colsep=2pt},
		rows={rowsep=1pt},
        hline{1,Z}={0.75pt},
		hline{2,3},
		vline{2}={2-Z}{solid},
		vline{4,7},
		cell{1}{4}={c=3}{c,m},
		cell{1}{7}={c=3}{c,m},
	}
	&&& Train time &&& Test time \\
	& N & D & DS-DGP & DSPP & AVDGP & DS-DGP & DSPP & AVDGP \\
	Wine & 6497 & 11
	& 2.57\,(0.02) & 3.27\,(0.03) & \tbf{2.41}\,(0.06) & 1.48\,(0.04) & 1.49\,(0.01) & \tbf{1.36}\,(0.01) \\
	Kin8nm & 8192 & 8
	& 3.45\,(0.01) & 4.34\,(0.02) & \tbf{3.17}\,(0.02) & 1.89\,(0.01) & 1.92\,(0.01) & \tbf{1.72}\,(0.01) \\
	Kin32nm & 8192 & 32
	& 3.49\,(0.03) & 4.37\,(0.01) & \tbf{3.16}\,(0.03) & 1.89\,(0.01) & 1.92\,(0.01) & \tbf{1.73}\,(0.02) \\
	WEC100 & 9595 & 200
	& 4.29\,(0.02) & 5.30\,(0.01) & \tbf{4.08}\,(0.02) & 2.21\,(0.01) & 2.25\,(0.01) & \tbf{2.12}\,(0.01) \\
	Elevator & 16599 & 18
	& 6.91\,(0.02) & 8.71\,(0.05) & \tbf{6.40}\,(0.07) & 3.59\,(0.01) & 3.67\,(0.04) & \tbf{3.37}\,(0.04) \\
	Bike & 17379 & 12
	& 7.32\,(0.04) & 9.21\,(0.03) & \tbf{6.70}\,(0.06) & 3.81\,(0.02) & 3.84\,(0.01) & \tbf{3.51}\,(0.02) \\
	Conductor &21263 & 81
	& 9.43\,(0.04) & 11.75\,(0.03) & \tbf{8.46}\,(0.06) & 4.71\,(0.01) & 4.78\,(0.02) & \tbf{4.29}\,(0.01) \\
	Protein & 45730 & 9
	& 19.38\,(0.05) & 24.38\,(0.06) & \tbf{17.80}\,(0.15) & 9.75\,(0.04) & 9.87\,(0.02) & \tbf{9.00}\,(0.04) \\
	KeggD & 53413 & 19
	& 22.98\,(0.15) & 28.70\,(0.04) & \tbf{21.16}\,(0.68) & 11.51\,(0.01) & 11.61\,(0.03) & \tbf{10.51}\,(0.02) \\
	CTSlice & 53500 & 379
	& \tbf{27.19}\,(0.04) & 33.12\,(0.09) & 28.40\,(0.08) & \tbf{12.76}\,(0.03) & 12.98\,(0.03) & 13.32\,(0.04) \\
	WEC49 & 54007 & 98
	& 24.40\,(0.05) & 30.49\,(0.06) & \tbf{21.95}\,(0.19) & 11.70\,(0.03) & 11.89\,(0.03) & \tbf{10.62}\,(0.04) \\
	KeggU & 64608 & 26
	& 27.68\,(0.05) & 34.91\,(0.04) & \tbf{25.24}\,(0.16) & 13.83\,(0.03) & 14.03\,(0.05) & \tbf{12.71}\,(0.05) \\
	Song & 515345 & 90
	& \tbf{61.88}\,(0.09) & 131.13\,(0.02) & 98.01\,(0.29) & 43.50\,(0.03) & 43.68\,(0.03) & \tbf{43.22}\,(0.04) \\
	Buzz & 583250 & 77
	& \tbf{69.62}\,(0.13) & 148.07\,(0.21) & 107.91\,(0.29) & 49.11\,(0.05) & 49.35\,(0.02) & \tbf{48.35}\,(0.06) \\
    \end{tblr}
	\label{tab4}
\end{table}

\begin{table} [ht]
	\footnotesize
	\centering
	\caption{
		A compilation of regression results from Sec. \ref{Sec_ar} and Sec. \ref{Sec_regr}.
		For each metric and dataset we bold the result for the best performing method
		(lower is better for all metrics). {\tpm} indicates standard deviations.
	}
	\begin{adjustbox}{minipage=20cm, center}
	\centering
	\begin{tblr}{
		cells={c,m},
		columns={colsep=2pt},
		rows={rowsep=1pt},
		hline{1,Z}={0.75pt},
		hline{3,17,31},
		vline{5},
		vline{9}={dotted},
		cell{1}{9}={c=4}{c,m},
	}
	&&&& SOLVE & DS-DGP & DSPP & IDSGP & AVDGP \\
	Metric & Dataset & N & D &&&&& AR1 & AR2 & AR2P & AR2PP \\
	NLL
	& Wine & 6497 & 11
	& 1.169\tpm0.020 & 1.190\tpm0.020 & 1.148\tpm0.021 & 1.180\tpm0.030
	& 1.163\tpm0.016 & 1.170\tpm0.025 & 1.193\tpm0.049 & \tbf{0.544}\tpm0.767 \\
	& Kin8nm & 8192 & 8
	& 0.236\tpm0.024 & 0.290\tpm0.024 & 0.147\tpm0.028 & 0.104\tpm0.022
	& 0.157\tpm0.014 & 0.171\tpm0.039 & \tbf{0.118}\tpm0.037 & \tbf{0.066}\tpm0.022 \\
	& Kin32nm & 8192 & 32
	& 1.147\tpm0.010 & 1.235\tpm0.061 & \tbf{1.095}\tpm0.016 & 1.143\tpm0.016
	& 1.112\tpm0.020 & 1.137\tpm0.019 & 1.120\tpm0.017 & \tbf{1.102}\tpm0.020 \\
	& WEC100 & 9595 & 200
	& 0.275\tpm0.016 & -0.099\tpm0.034 & -1.136\tpm0.038 & 0.013\tpm0.071
	& -1.659\tpm0.055 & \tbf{-2.031}\tpm0.035 & -1.955\tpm0.048 & -0.942\tpm0.172 \\
	& Elevator & 16599 & 18
	& 0.267\tpm0.009 & 0.108\tpm0.020 & \tbf{0.053}\tpm0.014 & 0.129\tpm0.009
	& 0.100\tpm0.014 & 0.117\tpm0.017 & 0.125\tpm0.021 & \tbf{0.054}\tpm0.012 \\
	& Bike & 17379 & 12
	& 0.890\tpm0.010 & -0.147\tpm0.023 & -0.568\tpm0.042 & -0.171\tpm0.036
	& -0.269\tpm0.026 & -0.270\tpm0.033 & -0.299\tpm0.035 & \tbf{-0.742}\tpm0.024 \\
	& Conductor & 21263 & 81
	& 0.548\tpm0.008 & 0.354\tpm0.009 & -0.272\tpm0.030 & -0.025\tpm0.020
	& -0.004\tpm0.017 & -0.040\tpm0.015 & -0.045\tpm0.015 & \tbf{-0.596}\tpm0.041 \\
	& Protein & 45730 & 9
	& 1.037\tpm0.009 & 1.036\tpm0.007 & 0.582\tpm0.029 & 0.994\tpm0.017
	& 0.957\tpm0.010 & 0.950\tpm0.015 & 0.910\tpm0.012 & \tbf{0.226}\tpm0.028 \\
	& KeggD & 53413 & 19
	& 0.133\tpm0.016 & -0.028\tpm0.013 & -1.551\tpm0.043 & -0.339\tpm0.043
	& & & -0.560\tpm0.030 & \tbf{-1.783}\tpm0.021 \\
	& CTSlice & 53500 & 379
	& -0.658\tpm0.005 & -2.014\tpm0.017 & -2.364\tpm0.020 & -3.085\tpm0.039
	& & & -3.236\tpm0.032 & \tbf{-3.312}\tpm0.040 \\
	& WEC49 & 54007 & 98
	& -0.384\tpm0.013 & -0.547\tpm0.011 & -2.049\tpm0.024 & -0.488\tpm0.035
	& & & -1.283\tpm0.009 & \tbf{-2.295}\tpm0.052 \\
	& KeggU & 64608 & 26
	& -0.068\tpm0.014 & -0.726\tpm0.032 & -2.970\tpm0.017 & -1.093\tpm0.080
	& & & -1.753\tpm0.034 & \tbf{-3.112}\tpm0.037 \\
	& Song & 515345 & 90
	& 1.177\tpm0.002 & 1.145\tpm0.003 & 0.859\tpm0.007 & 1.130\tpm0.003
	& & & 1.068\tpm0.021 & \tbf{0.651}\tpm0.005 \\
	& Buzz & 583250 & 77
	& -0.543\tpm0.015 & -0.673\tpm0.004 & -2.215\tpm0.005 & -0.664\tpm0.018
	& & & -0.727\tpm0.002 & \tbf{-2.239}\tpm0.005 \\
	RMSE
	& Wine & 6497 & 11
	& \tbf{0.787}\tpm0.016 & 0.803\tpm0.016 & \tbf{0.783}\tpm0.018 & \tbf{0.797}\tpm0.022
	& \tbf{0.786}\tpm0.015 & \tbf{0.785}\tpm0.021 & \tbf{0.793}\tpm0.028 & 0.808\tpm0.031 \\
	& Kin8nm & 8192 & 8
	& 0.302\tpm0.012 & 0.316\tpm0.006 & 0.283\tpm0.008 & 0.270\tpm0.006
	& 0.282\tpm0.003 & 0.279\tpm0.007 & \tbf{0.266}\tpm0.007 & \tbf{0.263}\tpm0.004 \\
	& Kin32nm & 8192 & 32
	& 0.760\tpm0.009 & 0.834\tpm0.048 & \tbf{0.728}\tpm0.008 & 0.757\tpm0.013
	& \tbf{0.735}\tpm0.016 & 0.749\tpm0.013 & 0.738\tpm0.013 & \tbf{0.727}\tpm0.015 \\
	& WEC100 & 9595 & 200
	& 0.331\tpm0.011 & 0.349\tpm0.018 & 0.304\tpm0.004 & 0.313\tpm0.014
	& \tbf{0.281}\tpm0.012 & \tbf{0.278}\tpm0.014 & \tbf{0.274}\tpm0.014 & 0.284\tpm0.014 \\
	& Elevator & 16599 & 18
	& 0.336\tpm0.010 & \tbf{0.289}\tpm0.003 & \tbf{0.290}\tpm0.005 & 0.293\tpm0.006
	& \tbf{0.288}\tpm0.005 & \tbf{0.288}\tpm0.005 & 0.295\tpm0.003 & 0.294\tpm0.003 \\
	& Bike & 17379 & 12
	& 0.591\tpm0.009 & 0.230\tpm0.009 & 0.251\tpm0.009 & 0.230\tpm0.010
	& 0.224\tpm0.007 & \tbf{0.216}\tpm0.006 & \tbf{0.218}\tpm0.006 & \tbf{0.216}\tpm0.005 \\
	& Conductor & 21263 & 81
	& 0.418\tpm0.009 & 0.363\tpm0.009 & 0.383\tpm0.011 & \tbf{0.285}\tpm0.013
	& \tbf{0.293}\tpm0.014 & \tbf{0.290}\tpm0.013 & \tbf{0.294}\tpm0.015 & 0.321\tpm0.015 \\
	& Protein & 45730 & 9
	& 0.683\tpm0.006 & 0.688\tpm0.005 & 0.679\tpm0.009 & 0.658\tpm0.009
	& 0.641\tpm0.007 & 0.635\tpm0.006 & \tbf{0.628}\tpm0.007 & 0.690\tpm0.006 \\
	& KeggD & 53413 & 19
	& 0.291\tpm0.005 & 0.252\tpm0.005 & 0.269\tpm0.007 & 0.214\tpm0.005
	& & & \tbf{0.204}\tpm0.004 & 0.285\tpm0.021 \\
	& CTSlice & 53500 & 379
	& 0.112\tpm0.005 & 0.035\tpm0.001 & 0.031\tpm0.002 & 0.014\tpm0.002
	& & & \tbf{0.011}\tpm0.001 & \tbf{0.011}\tpm0.001 \\
	& WEC49 & 54007 & 98
	& 0.201\tpm0.004 & 0.184\tpm0.005 & 0.196\tpm0.012 & 0.206\tpm0.004
	& & & \tbf{0.170}\tpm0.003 & 0.181\tpm0.007 \\
	& KeggU & 64608 & 26
	& 0.257\tpm0.010 & 0.182\tpm0.036 & 0.180\tpm0.010 & \tbf{0.133}\tpm0.006
	& & & \tbf{0.126}\tpm0.008 & 0.180\tpm0.046 \\
	& Song & 515345 & 90
	& 0.810\tpm0.001 & 0.793\tpm0.002 & 0.787\tpm0.001 & 0.781\tpm0.003
	& & & 0.812\tpm0.019 & \tbf{0.776}\tpm0.002 \\
	& Buzz & 583250 & 77
	& 0.280\tpm0.027 & \tbf{0.215}\tpm0.012 & 0.253\tpm0.027 & \tbf{0.220}\tpm0.029
	& & & \tbf{0.211}\tpm0.020 & 0.224\tpm0.006 \\
	CRPS
	& Wine & 6497 & 11
	& 0.501\tpm0.011 & 0.506\tpm0.011 & 0.488\tpm0.013 & 0.505\tpm0.013
	& 0.489\tpm0.009 & 0.494\tpm0.016 & 0.497\tpm0.017 & \tbf{0.461}\tpm0.026 \\
	& Kin8nm & 8192 & 8
	& 0.195\tpm0.005 & 0.195\tpm0.005 & 0.173\tpm0.005 & 0.172\tpm0.005
	& 0.174\tpm0.002 & 0.175\tpm0.004 & 0.168\tpm0.004 & \tbf{0.161}\tpm0.005 \\
	& Kin32nm & 8192 & 32
	& 0.496\tpm0.005 & 0.540\tpm0.033 & \tbf{0.458}\tpm0.008 & 0.494\tpm0.007
	& 0.472\tpm0.010 & 0.487\tpm0.010 & 0.477\tpm0.008 & \tbf{0.463}\tpm0.012 \\
	& WEC100 & 9595 & 200
	& 0.188\tpm0.003 & 0.153\tpm0.007 & 0.108\tpm0.003 & 0.157\tpm0.007
	& 0.105\tpm0.003 & \tbf{0.103}\tpm0.005 & \tbf{0.102}\tpm0.005 & \tbf{0.104}\tpm0.004 \\
	& Elevator & 16599 & 18
	& 0.206\tpm0.002 & 0.174\tpm0.002 & 0.170\tpm0.002 & 0.178\tpm0.001
	& 0.171\tpm0.002 & 0.175\tpm0.003 & 0.176\tpm0.002 & \tbf{0.161}\tpm0.004 \\
	& Bike & 17379 & 12
	& 0.356\tpm0.002 & 0.126\tpm0.004 & 0.125\tpm0.005 & 0.129\tpm0.006
	& 0.118\tpm0.003 & 0.116\tpm0.004 & 0.115\tpm0.003 & \tbf{0.094}\tpm0.003 \\
	& Conductor & 21263 & 81
	& 0.251\tpm0.003 & 0.208\tpm0.003 & 0.192\tpm0.006 & 0.150\tpm0.005
	& 0.152\tpm0.004 & 0.150\tpm0.003 & 0.150\tpm0.004 & \tbf{0.130}\tpm0.005 \\
	& Protein & 45730 & 9
	& 0.435\tpm0.005 & 0.423\tpm0.003 & 0.386\tpm0.006 & 0.402\tpm0.007
	& 0.379\tpm0.003 & 0.371\tpm0.006 & 0.357\tpm0.003 & \tbf{0.324}\tpm0.006 \\
	& KeggD & 53413 & 19
	& 0.162\tpm0.002 & 0.137\tpm0.002 & 0.112\tpm0.003 & 0.107\tpm0.003
	& & & \tbf{0.093}\tpm0.001 & 0.100\tpm0.005 \\
	& CTSlice & 53500 & 379
	& 0.070\tpm0.001 & 0.020\tpm0.000 & 0.015\tpm0.000 & 0.007\tpm0.000
	& & & 0.006\tpm0.000 & \tbf{0.005}\tpm0.000 \\
	& WEC49 & 54007 & 98
	& 0.101\tpm0.000 & 0.086\tpm0.001 & 0.062\tpm0.001 & 0.096\tpm0.003
	& & & 0.058\tpm0.001 & \tbf{0.046}\tpm0.002 \\
	& KeggU & 64608 & 26
	& 0.130\tpm0.002 & 0.067\tpm0.002 & 0.041\tpm0.002 & 0.053\tpm0.003
	& & & 0.036\tpm0.001 & \tbf{0.033}\tpm0.009 \\
	& Song & 515345 & 90
	& 0.486\tpm0.000 & 0.467\tpm0.001 & 0.439\tpm0.002 & 0.458\tpm0.002
	& & & 0.452\tpm0.003 & \tbf{0.359}\tpm0.001 \\
	& Buzz & 583250 & 77
	& 0.083\tpm0.001 & 0.074\tpm0.001 & 0.049\tpm0.001 & 0.075\tpm0.002
	& & & 0.071\tpm0.000 & \tbf{0.045}\tpm0.001 \\
	\end{tblr}
	\end{adjustbox}
	\label{tab5}
\end{table}

\section{Modified training objective in PPGPR} \label{App_pp}

\citet{pmlr-v119-jankowiak20a} proposed parametric predictive Gaussian process regression (PPGPR) whose
training objective directly targets the predictive distribution
\begin{equation*}
	p(\bsb{y}|\mbf{X},\mbf{Z})=\int p(\bsb{y}|\mbf{f})q(\mbf{f}|\mbf{X},\mbf{Z})\,d\mbf{f}
	=\mcal{N}\bigl(\bsb{y}|\ddot{\mbf{m}}_{\mbf{f}},\ddot{\mbf{K}}_{\mbf{ff}}+\sigma_{obs}^2\mbf{I}\bigr),
\end{equation*}
where the likelihood is Gaussian as in Eq. \ref{gauss_ll} and the marginal posterior over \(\mbf{f}\)
is given by Eq. \ref{sgp_marginal_posterior}.
They formulated a new objective function by modifying the expected log-likelihood term in the ELBO of SVGP
into logarithm of the marginal likelihood over targets \(\bsb{y}\)
and treating the KL term w.r.t. inducing variables as regularizer,
\begin{equation*}
	\mcal{L}_\text{ppgpr}=\ln\mcal{N}\bigl(
		\bsb{y}|\ddot{\mbf{m}}_{\mbf{f}},\ddot{\mbf{K}}_{\mbf{ff}}+\sigma_{obs}^2\mbf{I}
	\bigr)-\beta_\text{reg}D_\text{KL}\bigl(q(\mbf{u})\|p(\mbf{u}|\mbf{Z})\bigr),
\end{equation*}
where \(\beta_\text{reg}>0\) is an optional regularization constant.
Optimizing the above objective corresponds to finding maximum likelihood estimation
of a parametric model defined by the predictive distribution.
\citet{pmlr-v124-jankowiak20a} defined a similar objective for DSPP.

\section{CRPS for Gaussian mixtures} \label{App_crps}

The continuous ranked probability score (CRPS) for a continuous random variable
with cumulative distribution function (CDF) \(F\) is defined as
\begin{equation*}
	\text{CRPS}(F,x)=\int_{-\infty}^{\infty}\bigl[F(y)-\mbf{1}\{y\geq x\}\bigr]^2\,dy.
\end{equation*}
For a finite Gaussian mixture with CDF \(F=\sum_{s=1}^{S}\omega_s\mcal{N}(\mu_s,\sigma_s^2)\),
\citet{https://doi.org/10.1256/qj.05.235} derived an analytic expression for the defining integral as
\begin{equation*}
	\text{CRPS}(F,x)={\sum}_{s=1}^{S}\omega_sA(x-\mu_s,\sigma_s^2)
	-\frac{1}{2}{\sum}_{i=1}^{S}{\sum}_{j=1}^{S}\omega_i\omega_jA(\mu_i-\mu_j,\sigma_i^2+\sigma_j^2),
\end{equation*}
where
\begin{align*}
	A(\mu,\sigma^2) & =2\sigma\phi(\mu/\sigma)+\mu\bigl[2\Phi(\mu/\sigma)-1\bigr], \\
	\phi(x) & =\frac{1}{\sqrt{2\pi}}\exp\bigl\{-x^2/2\bigr\}, \\
	\Phi(x) & =\int_{-\infty}^{x}\phi(t)\,dt.
\end{align*}

\end{document}